\documentclass[journal]{IEEEtran}
\usepackage{ifpdf}

\usepackage{cite}

\ifCLASSINFOpdf
   \usepackage[pdftex]{graphicx}
\else
   \usepackage[dvips]{graphicx}
\fi
\usepackage{amsmath}
\usepackage{amssymb}
\usepackage{algorithm}
\usepackage{algorithmic}

\usepackage{array}

\ifCLASSOPTIONcompsoc
  \usepackage[caption=false,font=normalsize,labelfont=sf,textfont=sf]{subfig}
\else
  \usepackage[caption=false,font=footnotesize]{subfig}
\fi
\usepackage{dblfloatfix}
\usepackage{url}

\makeatletter
\let\NAT@parse\undefined
\makeatother
\usepackage{hyperref}  

\usepackage[T1]{fontenc}
\usepackage{bm}
\usepackage{color}
\usepackage{multirow}
\usepackage{multicol}
\usepackage{float}

\newcommand{\tabincell}[2]{\begin{tabular}{@{}#1@{}}#2\end{tabular}}

\begin{document}
%
\title{Modular Transfer Learning with Transition Mismatch Compensation for Excessive Disturbance Rejection}
%
%
%

\author{Tianming~Wang,~\IEEEmembership{Student~Member,~IEEE,}
        Wenjie~Lu,~\IEEEmembership{Member,~IEEE,}
        Huan~Yu,~\IEEEmembership{Student~Member,~IEEE,}
        and~Dikai~Liu,~\IEEEmembership{Member,~IEEE}
\thanks{T. Wang and D. Liu are with the Centre for Autonomous Systems, Faculty of Engineering and Information Technology, University of Technology Sydney, Ultimo, NSW 2007, Australia (e-mail: tianming.wang@student.uts.edu.au; dikai.liu@uts.edu.au).}
\thanks{W. Lu is with the School of Mechanical Engineering and Automation, Harbin Institute of Technology (Shenzhen), Shenzhen, 518055 Guangdong, P.R. China (e-mail: luwenjie@hit.edu.cn).}
\thanks{H. Yu is with the School of Automation, Beijing Institute of Technology, 100081 Beijing, P.R. China (e-mail: yuhuan.bit@gmail.com).}}

\maketitle

\begin{abstract}
Underwater robots in shallow waters usually suffer from strong wave forces, which may frequently exceed robot's control constraints. 
Learning-based controllers are suitable for disturbance rejection control, but the excessive disturbances heavily affect the state transition in Markov Decision Process (MDP) or Partially Observable Markov Decision Process (POMDP).
Also, pure learning procedures on targeted system may encounter damaging exploratory actions or unpredictable system variations, and training exclusively on a prior model usually cannot address model mismatch from the targeted system.
In this paper, we propose a transfer learning framework that adapts a control policy for excessive disturbance rejection of an underwater robot under dynamics model mismatch.
A modular network of learning policies is applied, composed of a Generalized Control Policy (GCP) and an Online Disturbance Identification Model (ODI). GCP is first trained over a wide array of disturbance waveforms. ODI then learns to use past states and actions of the system to predict the disturbance waveforms which are provided as input to GCP (along with the system state).
A transfer reinforcement learning algorithm using Transition Mismatch Compensation (TMC) is developed based on the modular architecture, that learns an additional compensatory policy through minimizing mismatch of transitions predicted by the two dynamics models of the source and target tasks.
We demonstrated on a pose regulation task in simulation that TMC is able to successfully reject the disturbances and stabilize the robot under an empirical model of the robot system, meanwhile improve sample efficiency. 
\end{abstract}


%
\IEEEpeerreviewmaketitle

\section{Introduction}
\label{section_introduction}


\IEEEPARstart{U}{nderwater} robotics has attracted an increasing interest from both research and industry in the last few decades. Recently, the applications of Autonomous Underwater Vehicle (AUV) and Remotely Operated Vehicle (ROV) to execute underwater tasks are more and more common, such as sea bottom survey, offshore structures monitoring, pipeline maintenance, biological samples collection and shipwreck search \cite{antonelli2018underwater,griffiths2002technology}. The rising demand for robotic advancements in all of these ocean research and industrial fields predicates the need to better understand the ocean dynamics.

Ocean waves will displace a robot during task execution. The wave forces decay exponentially from the water surface to the seabed, and sufficient depths yield negligible disturbances \cite{dean1991water}. Owing to this decay, as well as the considerable size and thrust capabilities of underwater robotic systems, the strength and changes of ocean waves are often neglected in robot motion planning and control in deep water applications \cite{fernandez2016model}.
In field applications with low operational depths and turbulent wave climates, like bridge pile inspection \cite{woolfrey2016kinematic} and sea-ice algae characterization in Antarctica \cite{wang2018case}, this assumption can quickly break down, since shallow water environments usually accommodate only small-size robots that have limited thrust capabilities, and the disturbances coming from the turbulent flows are time-varying and may frequently exceed robot's thrust capabilities (such wave forces are termed as excessive disturbances throughout this paper). As a result, increased wave forces inevitably hinder the stability and precision of robot motion control \cite{xie2000much,gao2014centrality,li2014disturbance}. 

\begin{figure}[t]
	\centering
	\includegraphics[width=\linewidth]{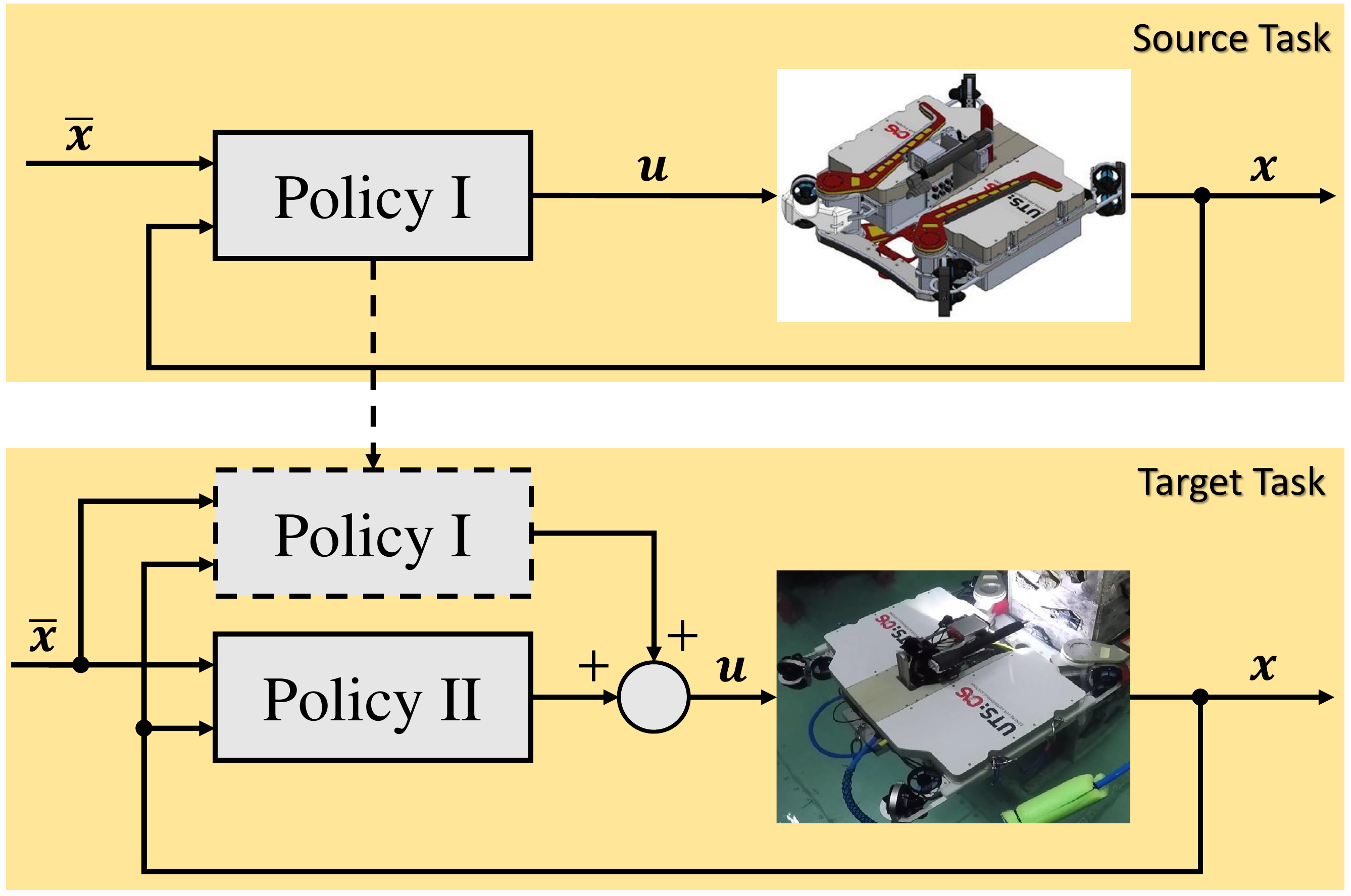}
	\caption{Diagram of transfer learning in reinforcement learning between two different dynamics models of underwater robots, Policy II produces a compensatory control action and runs in parallel with Policy I in the target task.}
	\label{fig_algorithm_diagram_transfer}
\end{figure}


%
Disturbance Observer Based Control (DOBC) \cite{chen2000nonlinear,chen2016disturbance,sun2016neural} has been widely investigated for decades. The main objective is to estimate the unknown disturbances from measurable variables, then take a control action based on the disturbance estimation to compensate for the influence of the disturbances.
%
In this method, external disturbances and model uncertainties are generally lumped together, then an observation mechanism is adopted to estimate the total disturbances. 
This leads to the change of the original properties for some disturbances, such as harmonic ones, thus making it difficult for predicting future disturbances.
%
A further problem is that naive addition of the disturbance estimation onto the control signal leads to inputs over constraints.

To this end, Model Predictive Control (MPC) \cite{garcia1989model,camacho2013model,fernandez2016model} is often applied due to its constraint handling capability through optimizing system behaviours over a future time horizon \cite{gao2016nonlinear}.
However, the performance of MPC relies on the accuracy of the given system dynamics model, and the requirement for online optimization at each timestep leads to low computational efficiency. 

In contrast, Reinforcement Learning (RL) \cite{sutton2018reinforcement} has drawn a lot of attention in learning optimal controllers for systems that are difficult to model accurately. It is a trial-and-error approach that allows to find an optimal sequence of commands without any prior assumption about the world, and can naturally adapt to uncertainties and noises (particularly noises that are independent and identically distributed) in the real system. Meanwhile, neural network controllers also enable high computational efficiency due to their fast forward propagation.
RL is superior in solving a Markov Decision Process (MDP), where the future states of the process depend only on the present state, not on the past states. But the excessive disturbances are not appropriate to be regarded as noises any more, since the state transition of the robot system is heavily affected by the unknown disturbances, leading to a violation of Markov property. 

Normally, when there are unobservable or hidden states to the agent, the problem will become a Partially Observable Markov Decision Process (POMDP). However, even if the disturbances are considered as the unobservable parts of the state space, the underlying transition function may still not be able to define a MDP, since it can be difficult to predict the next disturbances based on the current state (including disturbances) and action only.
Thus, the first research questions in this paper are how to find a better state space to represent the disturbances, so that the problem of excessive disturbance rejection can be formulated into a POMDP, then how to learn a control policy in such POMDP.

In addition, RL is able to successfully solve problems when being trained directly on the targeted system. Sometimes, training samples are expensive to obtain on the targeted system. For example, high sample complexity of deep RL algorithms often leads to long training time during their direct application to physical systems. Learning from scratch involves many exploratory actions, which real robots usually cannot withstand and may endanger both the agent and its surroundings. Other times, the targeted system may change over time in an unpredictable way during operation. These variations in the dynamics model may include variable friction coefficients, actuator failures, or varying load to be manipulated. In such cases, it is difficult to implement pure learning procedures on the targeted system.

A promising approach is to make an initial guess of the targeted system, such as a simulated counterpart of the real robot or a static model of the system before the operation, which is easy to get in most cases, then optimize a policy based on this model. However, because of the inaccurate replication of the targeted system dynamics, initially learned policies usually cannot be directly applied on the targeted system. Transfer learning offers a pathway to bridge the mismatch between different dynamics models.
Thus, the second research question in this paper is how to transfer a control policy for excessive disturbance rejection under dynamics model mismatch.


This paper introduces a transfer learning framework that enables successful deployment of a control policy for excessive disturbance rejection of an underwater robot under dynamics model mismatch, as shown in Fig.~\ref{fig_algorithm_diagram_transfer}. The contributions of this work are threefold.
Firstly, we seek a modular design of learning policies for excessive disturbance rejection, consisting of an observer network and a controller network. The observer is used to predict disturbance waveforms which are then provided as input to the controller together with the current system state to produce control action.
The modular architecture benefits policy transfer between different dynamics in sample efficiency, since only the controller network needs to be adapted, but the observer network can remain fixed.
Secondly, transfer learning is developed based on the modular architecture. To ensure the observer working correctly, an additional compensatory policy is learned through minimizing the mismatch of transitions predicted by the two dynamics models of the source and target tasks, respectively.
Such design enables the observer to predict the external disturbances in the target task. Then the model mismatch exists almost only in the internal dynamics, leading to reduced transfer difficulty.
Thirdly, the compensatory policy is added in terms of middle layer features instead of final network outputs in the target task, in order to offer more flexibility in compensation under the control constraints.

The transfer learning algorithms are evaluated on a pose regulation task under unobservable wave forces in simulation. In the evaluation, the source task defines a first-principle model of an underwater robot developed from the fundamental principles of dynamics, the target task applies an empirical model of an underwater robot derived from real-world experimental data. As a result, a control policy trained on the first-principle model is still able to successfully reject the disturbances and stabilize the robot on the empirical model through a small amount of adaptation.


In this paper, Section~\ref{section_review} covers a review of related work in the current literature. Section~\ref{section_formulation} introduces our problem formulation. Section~\ref{section_modular} provides the detailed description of the modular network design for excessive disturbance rejection control, followed by the details of the transfer learning algorithms in Section~\ref{section_transfer}. Then, Section~\ref{section_experiments} presents experimental evaluation procedures and result analysis. Limitations and some potential future improvements are discussed in the last section (Section~\ref{section_conclusion}).

\section{Related Work}
\label{section_review}

\subsection{Reinforcement Learning in Partially Observable Markov Decision Processes}

Deep RL algorithms based on Q-learning \cite{mnih2015human,oh2016control,gu2016continuous}, policy gradients \cite{schulman2015trust,gu2016q}, and actor-critic methods \cite{lillicrap2015continuous,mnih2016asynchronous,schulman2015high} have been shown to learn complex skills in high-dimensional state and action spaces, including video game playing, quadruped robot locomotion, autonomous driving, and dexterous manipulation.
However, real-world control problems rarely feature the full state information of the system. Then POMDP better describes the dynamics of real-world environments through explicitly recognizing that the agent only observes partial glimpses of the underlying system state.
Existing solutions to POMDP typically maintain a belief state over the world state given the observations so far. This method has shortcomings in model dependency and computational cost during belief update \cite{kaelbling1998planning,shani2013survey}.


Using Recurrent Neural Network (RNN) to represent policies is a more popular approach to handle partial observability \cite{wierstra2010recurrent,hausknecht2015deep,heess2015memory,mnih2016asynchronous,oh2016control}. The idea being that RNN is able to retain information from observations further back in time, and incorporate this information into predicting better actions and value functions, thus performing better on tasks that require long term planning. 
One of the earliest works is to apply RNN in Deep Q-Network (DQN) framework, which enables the policy to better handle POMDP by learning long-term dependencies. Hausknecht and Stone \cite{hausknecht2015deep} introduced a Deep Recurrent Q-Network (DRQN) that was capable of successfully estimating velocities in training video game player, where recurrent connections create an effective way to conditionally operate on previous observations that are far away in time.
Then attention mechanism was introduced for further improvements, leading to a Deep Attention Recurrent Q-Network (DARQN) \cite{sorokin2015deep}, that builds additional connections between recurrent units and lower layers. Attention mechanism allows the network to focus on the most important part of the next input, so that DARQN outperforms DRQN and DQN on the video games that require long-term planning.
In addition, RNN is not limited to the learning algorithms based on value function, there are also successful applications to policy gradients \cite{wierstra2010recurrent} and actor-critic methods \cite{heess2015memory,mnih2016asynchronous}.


The problem of excessive disturbance rejection has some differences from the generic POMDP. Normally, POMDP can be transformed back to MDP when there is full observability of the environment. However, this is not the case when the disturbances are considered as the unobservable parts of the state space, since it it difficult to formulate a transition function to predict next disturbances from current state (including disturbances) and action only.
Both history window approach \cite{wang2018excessive} and recurrent policy \cite{wang2019dob} attempt to resolve this issue through characterizing the disturbed system transition as a multi-step MDP, and assuming the unobservable disturbance waveforms are encoded in robot motion history. The difference lies in the way to use the history data, the history window approach directly takes most recent state-action pairs as additional input to the policy, while the recurrent policy employs RNN to explore past experience in order to learn an optimal embedding of history data.

In contrast to these "end-to-end" control policies, that take as input a sequence of motion history and directly output the optimal control, this paper explicitly decouples the process into disturbance identification and motion control. We believe that using the decoupled moderate-sized networks instead of a large network trained by RL in an end-to-end mode \cite{yu2017preparing}, might mitigate the learning difficulty and thus improve the sample efficiency, and such modular network design benefits policy transfer between different dynamics as well.

\subsection{Transfer Learning in Reinforcement Learning}

Although RL algorithms can learn complex skills in high-dimensional state and action spaces, it is almost impossible to directly deploy RL agent on real-world systems, due to the high sample complexity. In order to accelerate the learning process of RL, the knowledge previously obtained from related tasks can be used \cite{bengio2009curriculum,caruana1997multitask}. Transferring policy from one task to another \cite{taylor2009transfer}, especially from learning in physical simulators to adapting on real robots, has aroused great interest.
The most simple idea is to use identical network architecture for both simulation and real environment \cite{zhu2017target}. More sophisticated learning process adds new layers when transferring to the new task, in the meantime freezes old layers, thus avoids the problem of catastrophically forgetting \cite{rusu2016progressive,rusu2016sim}. There are also other methods including domain adaptation that learns aligned visual representations between synthetic and real-world images \cite{tzeng2020adapting,bousmalis2018using}.


Much of the previous work has focused on system identification \cite{lennart1999system,giri2010block}, which provides a framework for finding accurate system models, then simplifying the design of robot controllers in the real world.
In the context of RL, these methods are often referred to as model-based RL \cite{deisenroth2013survey}. That is, data from actual policy execution is used to fit a transition model and then used to learn a control policy without directly interacting with the real-world scene.
Such methods reduce the number of samples compared to purely model-free RL, which generally requires tremendous data to encode the objective function and the transition model.
Various model-based RL methods have been proposed \cite{deisenroth2011pilco,kuvayev1996model,forbes2002representations,hester2017intrinsically,jong2007model,sutton1991dyna}, and applied successfully on both simulated and real-world robots, such as inverted pendulums \cite{deisenroth2011pilco}, manipulators \cite{brauer2012using}, and legged robots \cite{schmidt2009distilling}. 

In fact, system identification is usually interleaved with policy optimization \cite{abbeel2005exploration,gevers2006system,bongard2005nonlinear,nagabandi2018neural}. In other words, additional policy execution data is collected through alternating between collecting data with the current model and retraining the model with the aggregated data. This data aggregation procedures improves performance by alleviating the mismatch between the distribution of the trajectory data and that of the model-based controller, and such an iterative process is able to converge to an optimal policy.


Another area of research is domain randomization, where the differences between the source and target tasks are modeled as the variabilities in the source task. Therefore, the source task can be designed to be as diverse as possible in the simulation in order to better generalize the trained policy to unfamiliar system dynamics or environmental factors in the real world.
Randomization in system dynamics have been used to design controllers that are robust to model uncertainties. Peng et al. \cite{peng2018sim} proposed to learn a policy in simulation through randomizing the physical parameters of the environment, then transfer the learned policy to a real robot for a puck pushing task. Andrychowicz et al. \cite{andrychowicz2020learning} proposed to train dexterous manipulation skills of a robotic hand using randomization of physical properties and object appearance in simulation.

However, sometimes it can be difficult for transfer learning using domain randomization, since this technique usually needs tedious manual fine-tuning and a significant expertise to design the distributions of simulation parameters \cite{chebotar2019closing}.
Thus, system identification and domain randomization have also been combined \cite{tan2018sim,lowrey2018reinforcement,antonova2017reinforcement}. These approaches address the problem of automatically learning the distributions of simulated parameters to avoid manually design procedures, thus improving the transfer learning of policies.
Rajeswaran et al. \cite{rajeswaran2016epopt} and Chebotar et al. \cite{chebotar2019closing} applied this framework to iteratively learn a policy over an ensemble model and used data from the target task to adjust model distributions. The results showed that policies can be successfully transferred with only a few iterations of simulation updates using a small number of real robot trials.


In summary, although neural-network-based dynamics model can make reasonable predictions over a future time horizon without physical interaction \cite{chiappa2017recurrent}, the success of model-based RL still heavily depends on the quality and quantity of the real-world experimental data. In order to promote the adoption of neural network models in model-based RL, finding strategies to improve their sample efficiency is necessary.
As for domain randomization, this kind of techniques generally restricted to only low-dimensional dynamics models. When the real-world dynamics become more complicated, the selected parameterization of the dynamics model might not well represent it. Thus domain randomization can be difficult to apply in sim-to-real transfer in most cases.

\section{Problem Formulation}
\label{section_formulation}

\subsection{Disturbed Robot Dynamics}


Consider a nonlinear time-invariant system to represent the dynamics of an underwater robot in the form of
\begin{equation}
	\label{eqn_complete_dynamics}
	\begin{gathered}
		\bm{M}\dot{\bm{\nu}}+\bm{C}\left(\bm{\nu}\right)\bm{\nu}+\bm{D}_{RB}\left(\bm{\nu}\right)\bm{\nu}+\bm{g}_{RB}\left(\bm{\eta}\right) = \bm{u}+\bm{d}+\bm{\xi}, \\
		\dot{\bm{\eta}} = \bm{J}\bm{\nu},
	\end{gathered}
\end{equation}
where $\bm{\eta},\bm{\nu} \in \mathbb{R}^{6}$ are the robot's pose and velocity, $\bm{x} = \left[\bm{\eta}^{T} \ \bm{\nu}^{T}\right]^{T} \in \mathcal{X} \in \mathbb{R}^{12}$ is the system state, $\mathcal{X}$ represents the state space,
$\bm{J} \in \mathbb{R}^{6\times6}$ is the system's Jacobian matrix, $\bm{M} \in \mathbb{R}^{6\times6}$ is the inertia matrix, $\bm{C}(\bm{\nu}) \in \mathbb{R}^{6\times6}$ is the matrix of Coriolis and centripetal terms, $\bm{M} = \bm{M}_{RB}+\bm{M}_{A}$ and $\bm{C} = \bm{C}_{RB}+\bm{C}_{A}$ include also added mass terms, $\bm{D}_{RB}(\bm{\nu}) \in \mathbb{R}^{6\times6}$ is the matrix of drag forces, $\bm{g}_{RB}(\bm{\eta}) \in \mathbb{R}^{6}$ is the vector of gravity and buoyancy forces,
$\bm{u} \in \mathcal{U} \in \mathbb{R}^{6}$ is the control vector applied to the system, $\mathcal{U}$ represents the action space, $\bm{d} \in \mathbb{R}^{6}$ represents the wave forces, and $\bm{\xi} \in \mathbb{R}^{6}$ represents the model uncertainties.

In this work, the wave forces $\bm{d}$ are considered as external disturbances to the system, all the other components in the system dynamics model \eqref{eqn_complete_dynamics} are termed as internal dynamics.
The external disturbances are assumed to be time-correlated signals following specific waveforms, and may frequently exceed the control constraints of the robot. 
We consider the control constraints of the form $\underline{\bm{u}} \le \bm{u} \le \overline{\bm{u}}$ with element-wise inequality, and $\underline{\bm{u}},\overline{\bm{u}} \in \mathbb{R}^{6}$ represent the respective lower and upper bounds. 
The model uncertainties $\bm{\xi}$ include parametric and structural uncertainties and exist in both internal dynamics and external disturbances. The sources of model uncertainties may include inaccurate estimation of hydrodynamics coefficients, control latency from unmodeled thruster dynamics, and so on.
%
In this work, we consider discretization of the continuous-time system in \eqref{eqn_complete_dynamics} modeled by
\begin{equation}
	\label{eqn_discrete_dynamics}
	\bm{x}_{t+1} = f\left(\bm{x}_{t},\bm{u}_{t}\right),
\end{equation}
which describes the evolution from time $t$ to $t+1$ of the state $\bm{x}$, given the action $\bm{u}$. 
Also notice that the parameters of the internal dynamics are assumed fixed but unknown to the learning algorithms.

\subsection{Control Objective}

In this control problem, the model of the dynamics $f$ is unknown to the controller.
A trajectory $\left(\bm{X},\bm{U}\right)$ is a sequence of controls $\bm{U} = \left\{\bm{u}_{0},\bm{u}_{1},\cdots,\bm{u}_{T-1}\right\}$, and corresponding state sequence $\bm{X} = \left\{\bm{x}_{0},\bm{x}_{1},\cdots,\bm{x}_{T-1},\bm{x}_{T}\right\}$ satisfying \eqref{eqn_discrete_dynamics}.
The objective function denoted by $J$ is the discounted sum of rewards $r$, incurred when the system starts from initial state $\bm{x}_{0}$ and is controlled by the control sequence $\bm{U}$ until the horizon $T$ is reached:
\begin{equation}
	\label{eqn_control_objective}
	J\left(\bm{x}_{0}\right) = \sum_{t=0}^{T-1}\gamma^{t}r\left(\bm{x}_{t},\bm{u}_{t}\right),
\end{equation}
where the reward function is given as:
\begin{equation}
	\label{eqn_task_reward}
	r\left(\bm{x}_{t},\bm{u}_{t}\right) = -\bm{x}_{t}^{T}\bm{Q}\bm{x}_{t}-\bm{u}_{t}^{T}\bm{R}\bm{u}_{t},
\end{equation}
with $\bm{Q} \in \mathbb{R}^{n_{x} \times n_{x}}$ and $\bm{R} \in \mathbb{R}^{n_{u} \times n_{u}}$ being weight matrices, and $\gamma \in [0,1]$ is a discount factor that prioritizes near-term rewards.
The trajectory is implicitly represented using only the controls $\bm{U}$. The state sequence $\bm{X}$ is recovered by interaction with the environment \eqref{eqn_discrete_dynamics} from the initial state $\bm{x}_{0}$. 
%

In this work, RL is used to optimize this control objective through a trial-and-error approach. At each timestep $t$, the system makes a transition from the state variable $\bm{x}_{t}$ to $\bm{x}_{t+1}$ in response to a control signal $\bm{u}_{t}$ chosen from some policy $\pi$ under a dynamics model $f$, meanwhile collecting a scalar reward $r_{t}$ according to a reward function $r$. 
The goal of RL is to learn a policy $\bm{u}_{t} = \pi\left(\bm{x}_{t}\right)+\bm{n}, \bm{n} \sim \mathcal{N}$ that maximizes the objective function $J$:
\begin{equation}
	\label{eqn_optimization}
	\pi^{*} = \underset{\pi}{\operatorname{argmax}} J\left(\bm{x}_{0}\right).
\end{equation}

\subsection{Transfer Learning}

The key idea of transfer learning \cite{taylor2009transfer} is that experience or knowledge obtained from learning to specialize in one task can help improve learning effectiveness in a different but related task. The former is called source task, the latter is called target task. In general, task and MDP are used interchangeably.

In this research, the source and target tasks differ in system dynamics. The source task defines a mathematical model of an underwater robot developed from the fundamental principles of dynamics, this model corresponds to the dynamics function in \eqref{eqn_complete_dynamics} excluding the model uncertainties term $\bm{\xi}$, and is referred to as "first-principle model". The target task applies a data-driven model of an underwater robot derived from real-world experimental data, this model contains various uncertainties in the real system, and is referred to as "empirical model".
Specifically, the empirical model consists of two parts, the first part is a deep neural network learned from real-time motion data of the robot system in still water, representing internal dynamics; the other part is force data of real-world ocean waves collected in open water, representing external disturbances.
These two models differ in both internal dynamics and external disturbances.
But in the meantime, the state and action spaces, the reward and objective functions between the source and target tasks are all kept consistent.
Furthermore, the information transferred between the tasks is the control policy. 

In this paper, the policy directly trained on the source task is defined as "source policy"; the policy directly trained on the target task is defined as "target policy"; the policy trained on the source task then directly applied on the target task without transfer learning is defined as "unadapted policy"; the policy trained on the source task then further adapted on the target task using transfer learning is defined as "adapted policy".

There are many metrics to measure the benefits of transfer. In this work, we use jumpstart and learning time. That is, the goals of transfer learning are to improve the initial performance of an agent in a target task, and to reduce the learning time required by the agent to achieve optimal performance, compared with learning from scratch in the target task.
%
Learning time is regarded as a surrogate for sample complexity, which refers to the amount of data required by a learning algorithm to converge. These two concepts are strongly correlated, because RL agents collect data merely through repeated interactions with an environment.

\section{Modular Network Design}
\label{section_modular}

\begin{figure}[t]
	\centering
	\includegraphics[width=\linewidth]{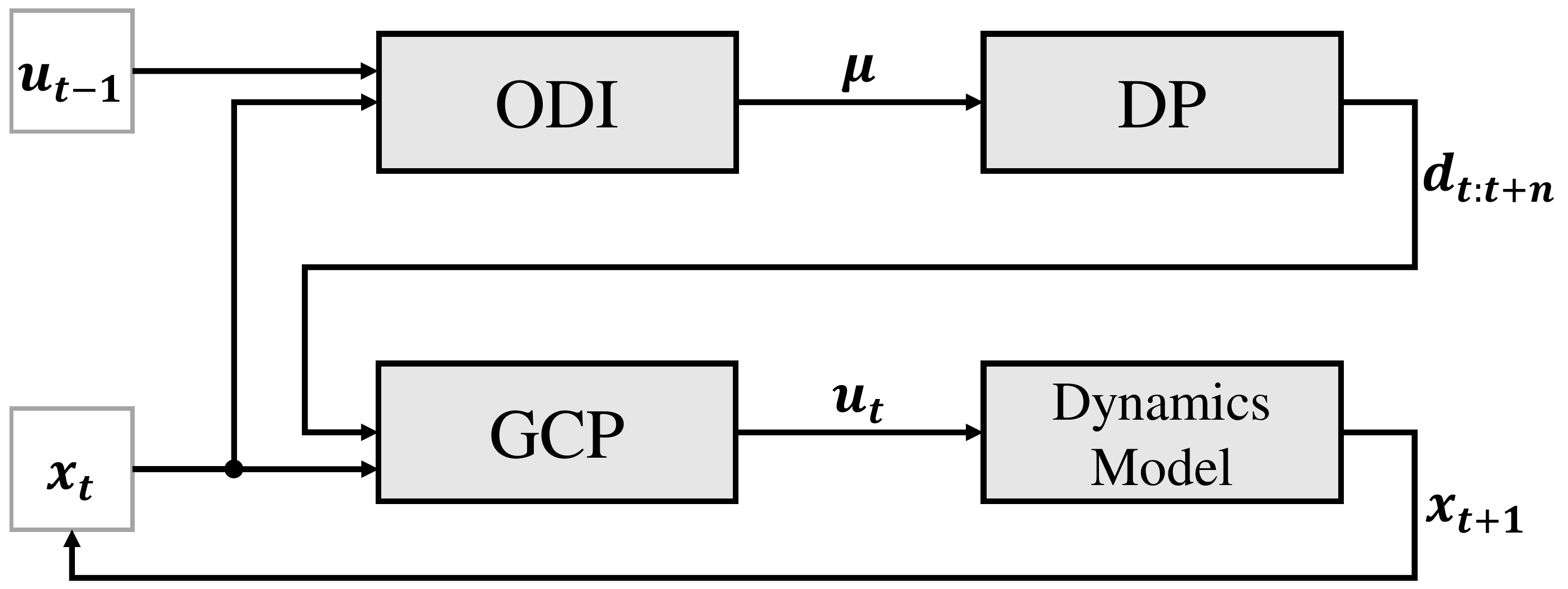}
	\caption{Diagram of GCP-ODI. The Online Disturbance Identification Model (ODI) employs RNN to identify the disturbance parameters $\bm{\mu} = \left\{A_{1},\omega_{1},\phi_{1},\cdots,A_{k},\omega_{k},\phi_{k}\right\}$ from the past states and actions, and the Disturbance Prediction Model (DP) formulates a sequence of future disturbances $\bm{d}_{t:t+n}$ from these parameters, where $\bm{d}_{t} = A_{1}\sin\left(\omega_{1}t+\phi_{1}\right) + \cdots + A_{k}\sin\left(\omega_{k}t+\phi_{k}\right)$. The Generalized Control Policy (GCP) then takes the predicted future disturbances $\bm{d}_{t:t+n}$ along with the current state $\bm{x}_{t}$ to compute the control action $\bm{u}_{t}$.}
	\label{fig_algorithm_diagram_gcp_odi}
\end{figure}

Previous work \cite{wang2019dob} has demonstrated that RNN can directly learn to control a dynamical system with unobservable disturbances in an end-to-end mode, where the past motion history is mapped to the control action. While, inspired by \cite{yu2017preparing}, this work applies modular learning procedures, that explicitly decouple the process into disturbance identification and motion control.

The learning algorithm for excessive disturbance rejection control proposed in this paper is composed of two main modules, namely a Generalized Control Policy (GCP) and an Online Disturbance Identification Model (ODI), as shown in Fig.~\ref{fig_algorithm_diagram_gcp_odi}. 
Firstly, we build a RL framework to train GCP, $\bm{u}_{t} = \pi\left(\bm{x}_{t},\bm{d}_{t:t+n}\right)$, under a system dynamics model, $\bm{x}_{t+1} = f\left(\bm{x}_{t},\bm{u}_{t};\bm{\mu}\right)$, parameterized using the disturbance parameters $\bm{\mu}$. Unlike classical RL policies where a mapping between states and controls is established, GCP explicitly takes a sequence of future disturbances (i.e. $\bm{d}_{t:t+n}$) as input besides the current state, and outputs a control signal. This additional input allows the policy to specialize at each set of disturbance waveforms, and is shown to improve the control performance in Section~\ref{section_experiments}. 
Secondly, we employ RNN to construct ODI, $\left(\bm{\mu},\bm{h}_{t}\right) = \psi\left(\bm{x}_{t},\bm{u}_{t-1},\bm{h}_{t-1}\right)$, and train it following a supervised learning style. 
%

When combining GCP and ODI together, the complete workflow is to use ODI to predict the disturbance parameters $\bm{\mu}$ based on the current system state $\bm{x}_{t}$, the previously executed action $\bm{u}_{t-1}$ of the robot, and the hidden state vector $\bm{h}_{t-1}$ from the previous timestep as well. The Disturbance Prediction Model (DP) uses these parameters to produce a sequence of future disturbances $\bm{d}_{t:t+n}$, which are then fed into GCP, with the current system state $\bm{x}_{t}$, to generate the control action $\bm{u}_{t}$. The action $\bm{u}_{t}$ is executed on the robot and the system dynamics gives an updated state $\bm{x}_{t+1}$, then the algorithm proceeds to the next timestep (Fig.~\ref{fig_algorithm_diagram_gcp_odi}).

We speculate that using the decoupled moderate-sized networks instead of a large network trained by RL in an end-to-end mode, might mitigate the learning difficulty and thus improve the sample efficiency. The modular learning procedures can also benefit the policy transfer under dynamics model mismatch, since only the controller network needs to be modified for adaptation to the target task, the identification network can remain fixed, thus reducing the transfer difficulty.

Both GCP and ODI are parameterized as deep neural networks. During training, GCP is trained first to cover all of the possible disturbance waveforms that ODI might explore during following optimization, where there is no need for any initial network or prior knowledge for both modules.

\subsection{Learning Generalized Control Policy}

\begin{algorithm}[t]
	\caption{Learning Generalized Control Policy} 
	\label{alg_gcp}	
	\begin{algorithmic}[1]
		\STATE Randomly initialize policy network $\pi$
		\STATE Initialize rollout buffer $R$
		\STATE Initialize episode step counter $t = 0$
		\STATE Sample state $\bm{x}_{t} \sim p\left(\bm{x}_{0}\right)$
		\STATE Sample disturbance parameters $\bm{\mu} \sim p\left(\bm{\mu}\right)$
		\STATE Initialize disturbance sequence $\bm{d}_{t:t+n}$ from $\bm{\mu}$
		\STATE Insert $\left(\bm{x}_{t},\bm{d}_{t:t+n}\right)$ into $R$
		\WHILE{$step \leq MaxStep$} 
			\WHILE{$R$ is not full}
				\STATE Calculate disturbances $\bm{d}_{t}$ from $\bm{\mu}$
				\STATE Perform action $\bm{u}_{t} = \pi\left(\bm{x}_{t},\bm{d}_{t:t+n}\right)$
				\STATE Receive next state $\bm{x}_{t+1} = f\left(\bm{x}_{t},\bm{u}_{t};\bm{\mu}\right)$
				\STATE Receive reward $r_{t} = r\left(\bm{x}_{t},\bm{u}_{t}\right)$
				\STATE Update $\bm{d}_{t+1:t+1+n}$
				\STATE Insert $\left(\bm{x}_{t+1},\bm{d}_{t+1:t+1+n},\bm{u}_{t},r_{t}\right)$ into $R$
				\STATE Update $t \leftarrow t+1$
				\IF{episode is terminated}
					\STATE Reset $t = 0$
					\STATE Sample $\bm{x}_{t} \sim p\left(\bm{x}_{0}\right)$
					\STATE Sample $\bm{\mu} \sim p\left(\bm{\mu}\right)$
				\ENDIF
			\ENDWHILE
			\STATE Update $\pi$ using data in $R$
		\ENDWHILE
	\end{algorithmic}
\end{algorithm}

The disturbance rejection controller is expected to be generalized over a range of disturbance waveforms. The intuition for this problem is to directly train one unified control policy that adapts to all different waveforms. While our initial attempt showed that, the policies change a lot when optimized over different waveforms, and the trained unified policy can hardly succeed at the given task. The reason behind is that the wave forces are too strong to be considered as random noises any more, and the disturbed system dynamics under different waveforms tends to be drastically varied with each other. Thus, training one single policy to deal with a wide variety of disturbances usually leads to poor generalization.

In this work, we found that it is possible to train a "generalized" policy to specialize at each set of waveforms through constructing the policy network parameterized by $\bm{\mu}$. After sufficient training, the generalized policy can achieve high cumulative rewards over the space of $\bm{\mu}$, 
%
its performance can be comparable with the policies trained for a specific set of waveforms based on state input only.
However, the disturbance parameters $\bm{\mu}$ are composed of frequency domain signals $\left\{A_{1},\omega_{1},\phi_{1},\cdots,A_{k},\omega_{k},\phi_{k}\right\}$, simply appending these signals to the input state cannot yield a well-performed control policy (see Section~\ref{section_experiments}). 
The reason behind this phenomenon is that the frequency domain signals only provide a description of the waveforms, but do not explicitly indicate phase information, so that the algorithm cannot figure out where it is along the waveforms at each timestep. 
One improvement is to formulate time domain signals from the frequency domain signals, and use multi-step future disturbances $\bm{d}_{t:t+n}$ as additional input for the policy, where $\bm{d}_{t} = A_{1}\sin\left(\omega_{1}t+\phi_{1}\right) + \cdots + A_{k}\sin\left(\omega_{k}t+\phi_{k}\right)$. 
%
Compared with using the frequency domain signals, the time domain signals do not contain complete information of the disturbances. But they directly give the disturbance waveforms in the near future, which are more closely related to control. Thus, they are still proved to achieve better performance in Section~\ref{section_experiments}.

Advantage Actor Critic (A2C) \cite{mnih2016asynchronous} is used to train GCP. During each episode, the algorithm (Algorithm~\ref{alg_gcp}) samples disturbance parameters $\bm{\mu}$ from a uniform distribution $p\left(\bm{\mu}\right)$, and constructs a sequence of future disturbances in the time domain $\bm{d}_{t:t+n}$, then performs the trajectory under the policy $\pi\left(\bm{x}_{t},\bm{d}_{t:t+n}\right)$ and the dynamics model $f\left(\bm{x}_{t},\bm{u}_{t};\bm{\mu}\right)$. Once the trajectory data are collected, the policy network $\pi$ will be updated following A2C rules.

\subsection{Learning Online Disturbance Identification Model}

\begin{algorithm}[t]
	\caption{Learning Online Disturbance Identification Model} 
	\label{alg_odi}	
	\begin{algorithmic}[1]
		\STATE Randomly initialize ODI netowrk $\psi$
		\STATE Initialize training buffer $B$
		\STATE Initialize iteration counter $iter = 1$
		\WHILE{$\psi$ is not converged}
			\FOR{$i = 1:K$}
				\STATE Sample $\bar{\bm{\mu}} \sim p\left(\bm{\mu}\right)$
				\FOR{$j = 1:N$}
					\STATE Sample $\bm{x}_{0} \sim p\left(\bm{x}_{0}\right)$
					\FOR{$t = 0:T-1$}
						\STATE Calculate $\bar{\bm{d}}_{t}$ from $\bar{\bm{\mu}}$
						\IF{$iter == 1$}
							\STATE Calculate $\bar{\bm{d}}_{t:t+n}$ from $\bar{\bm{\mu}}$
							\STATE Perform $\bm{u}_{t} = \pi\left(\bm{x}_{t},\bar{\bm{d}}_{t:t+n}\right)$
						\ELSIF{$iter > 1$}
							\STATE Predict $\left(\hat{\bm{\mu}},\bm{h}_{t}\right) = \psi\left(\bm{x}_{t},\bm{u}_{t-1},\bm{h}_{t-1}\right)$
							\STATE Calculate $\hat{\bm{d}}_{t:t+n}$ from $\hat{\bm{\mu}}$
							\STATE Perform $\bm{u}_{t} = \pi(\bm{x}_{t},\hat{\bm{d}}_{t:t+n})$
						\ENDIF
						\STATE Receive $\bm{x}_{t+1} = f\left(\bm{x}_{t},\bm{u}_{t};\bar{\bm{\mu}}\right)$
					\ENDFOR
					\STATE Form trajectory $\bm{\tau} = \left(\bm{u}_{0},\bm{x}_{1},\cdots,\bm{u}_{T-1},\bm{x}_{T}\right)$
					\STATE Store $\left(\bm{\tau},\bar{\bm{\mu}}\right)$ in $B$ \\
				\ENDFOR
			\ENDFOR
			\STATE Optimize $\psi$ using data in $B$ \\
			\STATE Update $iter \leftarrow iter+1$ \\
		\ENDWHILE
	\end{algorithmic}
\end{algorithm}

Even though GCP is capable of performing optimal control under different waveforms, the policy can only succeed with the correct disturbance parameters, but this information is usually not easy to obtain. This issue can be addressed by learning an Online Disturbance Identification Model (ODI), $\bm{\mu} = \psi\left(\bm{x}_{t},\bm{u}_{t-1},\bm{h}_{t-1}\right)$, that continuously identifies the correct disturbance parameters for GCP.

The training of ODI can be framed as a supervised learning problem, where the model aims to predict the disturbance parameters $\bm{\mu}$ with the input being the most recent state-action pair $\left(\bm{x}_{t},\bm{u}_{t-1}\right)$ and the hidden state $\bm{h}_{t-1}$ from the previous timestep. The optimization is conducted through minimizing the objective function using Stochastic Gradient Descent (SGD):
\begin{equation}
	\label{eqn_odi_objective}
	\theta_{\psi}^{*} = \underset{\theta_{\psi}}{\operatorname{argmin}} \sum_{\left(\bm{\tau}_{i},\bm{\mu}_{i}\right) \in B} \sum_{t=0}^{T-1} \left\|\bm{\mu}_{i}-\psi\left(\bm{x}_{t},\bm{u}_{t-1},\bm{h}_{t-1};\theta_{\psi}\right)\right\|^{2},
\end{equation}
where $\bm{\tau} = \left\{\bm{x}_{0},\bm{u}_{0},\cdots,\bm{x}_{T-1},\bm{u}_{T-1},\bm{x}_{T}\right\}$ is a trajectory of states and actions in each episode, and $\theta_{\psi}$ represents the parameters of ODI network $\psi$. During training, we randomly sample the space of $\bm{\mu}$ for $K$ times, and we simulate $N$ episodes for each sampled $\bar{\bm{\mu}}$, with the policy $\pi\left(\bm{x}_{t},\bar{\bm{d}}_{t:t+n}\right)$ and the dynamics $f\left(\bm{x}_{t},\bm{u}_{t};\bar{\bm{\mu}}\right)$. The trajectory data is then stored in the training buffer $B$ and used to update ODI network $\psi$ using \eqref{eqn_odi_objective} (Algorithm~\ref{alg_odi}).

After optimizing ODI $\psi$ using \eqref{eqn_odi_objective}, we noticed that the performance of the combined algorithm, GCP-ODI, was much worse than directly feeding the true disturbance parameters $\bar{\bm{\mu}}$ to GCP (see Section~\ref{section_experiments}). This result is not surprising since the trajectories used for training ODI are generated by performing a control policy $\pi\left(\bm{x}_{t},\bar{\bm{d}}_{t:t+n}\right)$ under a dynamics model $f\left(\bm{x}_{t},\bm{u}_{t};\bar{\bm{\mu}}\right)$, where they both use the same disturbance parameters $\bar{\bm{\mu}}$ in their formulation. However, when ODI is deployed with GCP, due to the lack of information at the initial timestep and the regression error in the neural network, ODI will inevitably make some error in the prediction. This error tends to be more and more aggravated over time because the next state-action pair fed to ODI is generated by the control policy $\pi(\bm{x}_{t},\hat{\bm{d}}_{t:t+n})$ using an incorrectly predicted disturbance parameters $\hat{\bm{\mu}}$, under the dynamics model with the true disturbance parameters $f\left(\bm{x}_{t},\bm{u}_{t};\bar{\bm{\mu}}\right)$.

In order to solve this issue, one possible idea is to iteratively train ODI by using mismatched disturbance parameters ($\hat{\bm{\mu}} \neq \bar{\bm{\mu}}$) for the control policy $\pi(\bm{x}_{t},\hat{\bm{d}}_{t:t+n})$ and the system dynamics $f\left(\bm{x}_{t},\bm{u}_{t};\bar{\bm{\mu}}\right)$. In each iteration, more training samples are generated following the previous procedures, but the disturbances used by GCP now come from the predicted disturbance parameters of ODI $\hat{\bm{\mu}}$, rather than the true disturbance parameters $\bar{\bm{\mu}}$ used in the system dynamics. Then, ODI is trained again through combining the mismatched training samples with previously gathered ones according to \eqref{eqn_odi_objective}. After a small number of iterations, the combined system, GCP-ODI, achieves close performance with GCP that is fed with the true disturbance parameters $\bar{\bm{\mu}}$.

\section{Transfer Learning Algorithm}
\label{section_transfer}

Previous work on transfer learning in RL \cite{peng2018sim,chebotar2019closing,koryakovskiy2018model} generally consider model discrepancies as a whole. Then in the problem of excessive disturbance rejection, the model mismatch in both external disturbances and internal dynamics will be combined together and treated equally by the existing transfer learning algorithms, without taking full advantage of the characteristics of the underwater disturbances.
While in this work, the external disturbances are separated from the internal dynamics and processed independently, through using ODI to predict a set of disturbance waveforms that best fit the real ones in the target task. The residual mismatch in the external disturbances, which is shown to be sufficiently small in Section~\ref{section_experiments}, is then merged into the mismatch of the internal dynamics and adapted together by the transfer learning. Such framework is expected to reduce the amount of total model mismatch that the transfer process needs to adapt, thus improves the learning efficiency during transfer.

This section discusses three transfer RL algorithms, Control Action Compensation (CAC), Transition Mismatch Compensation with Control Combination (TMC-control) and with Feature Combination (TMC-feature), based on the modular architecture of GCP-ODI proposed in Section~\ref{section_modular}. All of them follow a basic idea of training an additional policy to compensate the dynamics model mismatch between the source and target tasks, as shown in Fig.~\ref{fig_algorithm_diagram_transfer}.

\subsection{Control Action Compensation}

\begin{figure}[t]
	\centering
	\includegraphics[width=\linewidth]{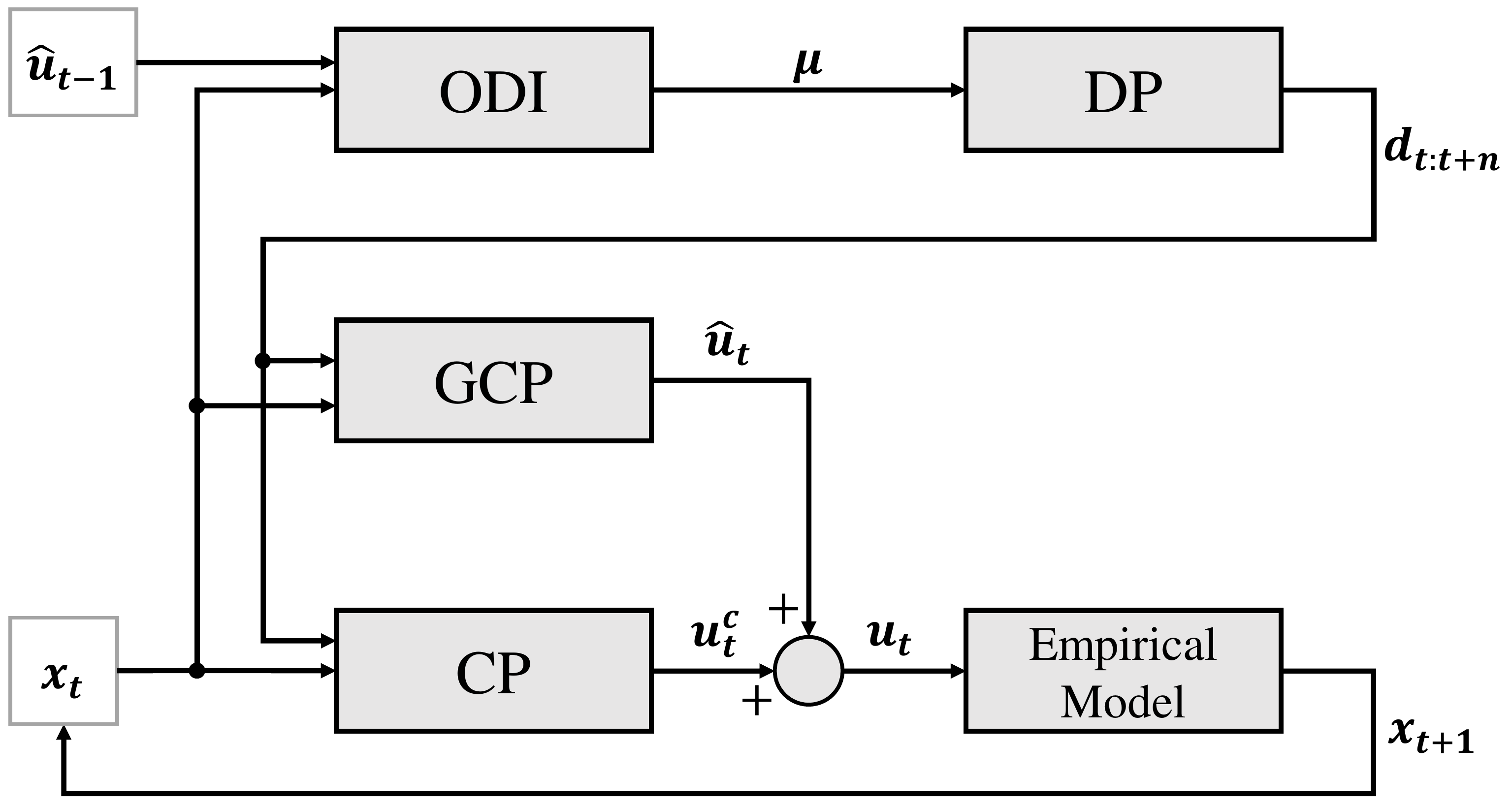}
	\caption{Diagram of transfer reinforcement learning using Control Action Compensation (CAC).}
	\label{fig_algorithm_diagram_cac}
\end{figure}

We first propose a transfer RL algorithm using Control Action Compensation (CAC), as shown in Fig.~\ref{fig_algorithm_diagram_cac}. 
This algorithm learns an additional control policy, which generates a compensatory control action $\bm{u}^{c}_{t}$ directly added to the control input computed by the source policy (i.e. GCP) $\hat{\bm{u}}_{t}$, the combined control action $\bm{u}_{t}$ is then applied to the empirical model in the target task. 
The training process uses the task reward function \eqref{eqn_task_reward}, which establishes similar optimization goals for both the source policy and the compensatory policy.
In order to ensure that ODI remains effective, the action fed into ODI should be the output of the source policy $\hat{\bm{u}}_{t}$, that is what ODI has seen during training. However, through transfer learning using \eqref{eqn_task_reward}, CAC algorithm could not optimize the compensatory policy to a satisfactory performance. This is because, although ODI is fed with the action of the source policy, the whole trajectory data still comes from the combined control policy and the empirical model in the target task. In contrast, ODI is trained using the trajectory data generated by the source policy under the first-principle model in the source task. Thus, the differently distributed trajectories fed to ODI lead to worse performance of the whole system.

\subsection{Transition Mismatch Compensation}

\begin{figure}[t]
	\centering
	\includegraphics[width=\linewidth]{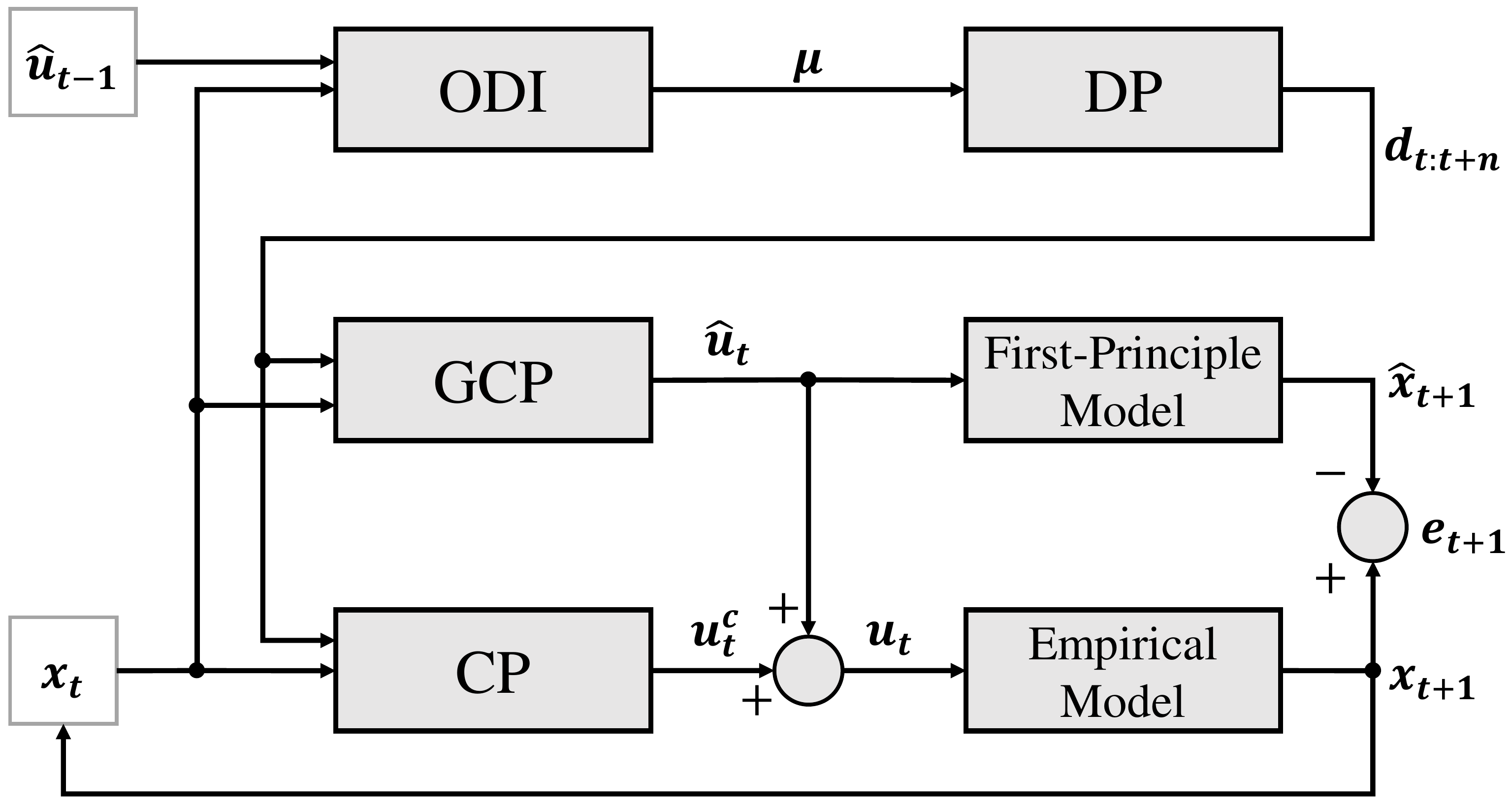}
	\caption{Diagram of transfer reinforcement learning using Transition Mismatch Compensation with Control Combination (TMC-control).}
	\label{fig_algorithm_diagram_tmc_control}
\end{figure}

To address the issue of trajectory mismatch, we develop another transfer RL algorithm based on Transition Mismatch Compensation (TMC), as shown in Fig.~\ref{fig_algorithm_diagram_tmc_control}. 
This algorithm learns a compensatory policy from the difference between transitions of the first-principle model and the empirical model, $\bm{e}_{t} = \bm{x}_{t}-\hat{\bm{x}}_{t}$. 
In this case, the compensatory policy uses a different optimization goal where the reward is given by
\begin{equation}
	\label{eqn_mismatch_reward}
	\begin{split}
		r_{m}\left(\bm{x}_{t},\bm{u}_{t},\bm{x}_{t+1}\right) &= r\left(\bm{x}_{t},\bm{u}_{t}\right) - \left\|\bm{e}_{t+1}\right\|_{2} \\
		&= -\bm{x}_{t}^{T}\bm{Q}\bm{x}_{t} - \bm{u}_{t}^{T}\bm{R}\bm{u}_{t} - \left\|\bm{x}_{t+1}-\hat{\bm{x}}_{t+1}\right\|_{2}
	\end{split},
\end{equation}
which does not divert the source policy from reaching its objective. The purpose of such optimization setting is to eliminate the transition mismatch $\bm{e}$ by forcing the empirical model to behave like the first-principle model as if there is no model mismatch. When the transition mismatch approaches zero $\bm{e}_{t} \rightarrow 0$, the state trajectories of the empirical model will get similar with those of the first-principle model, making ODI work correctly by using the trajectory data matched with the training process. In the meantime, the outcome of the combined control policy will approach the outcome of the optimal policy with respect to the first-principle model. Also, under this setting, the well trained ODI is able to predict a set of waveforms that well describe the real disturbances from the distributions of simulated parameters, thus the total model mismatch required to be compensated is reduced.

However, these optimization effects are achievable only when the full power of the compensatory control is released. We found that, in order to have good performance against the excessive disturbances, the control actions of the source policy $\hat{\bm{u}}_{t}$ often reach or exceed the control constraints, leaving a little space for the compensation to directly take effect. Thus, the compensatory term needs to learn a very complicated function in order to optimize overall system performance under the constraints of robot control capabilities. 
%
That is why we believe the transfer learning has not reached its performance limit, and we may need a more efficient way to apply the control compensation.

\begin{figure}[t]
	\centering
	\includegraphics[width=\linewidth]{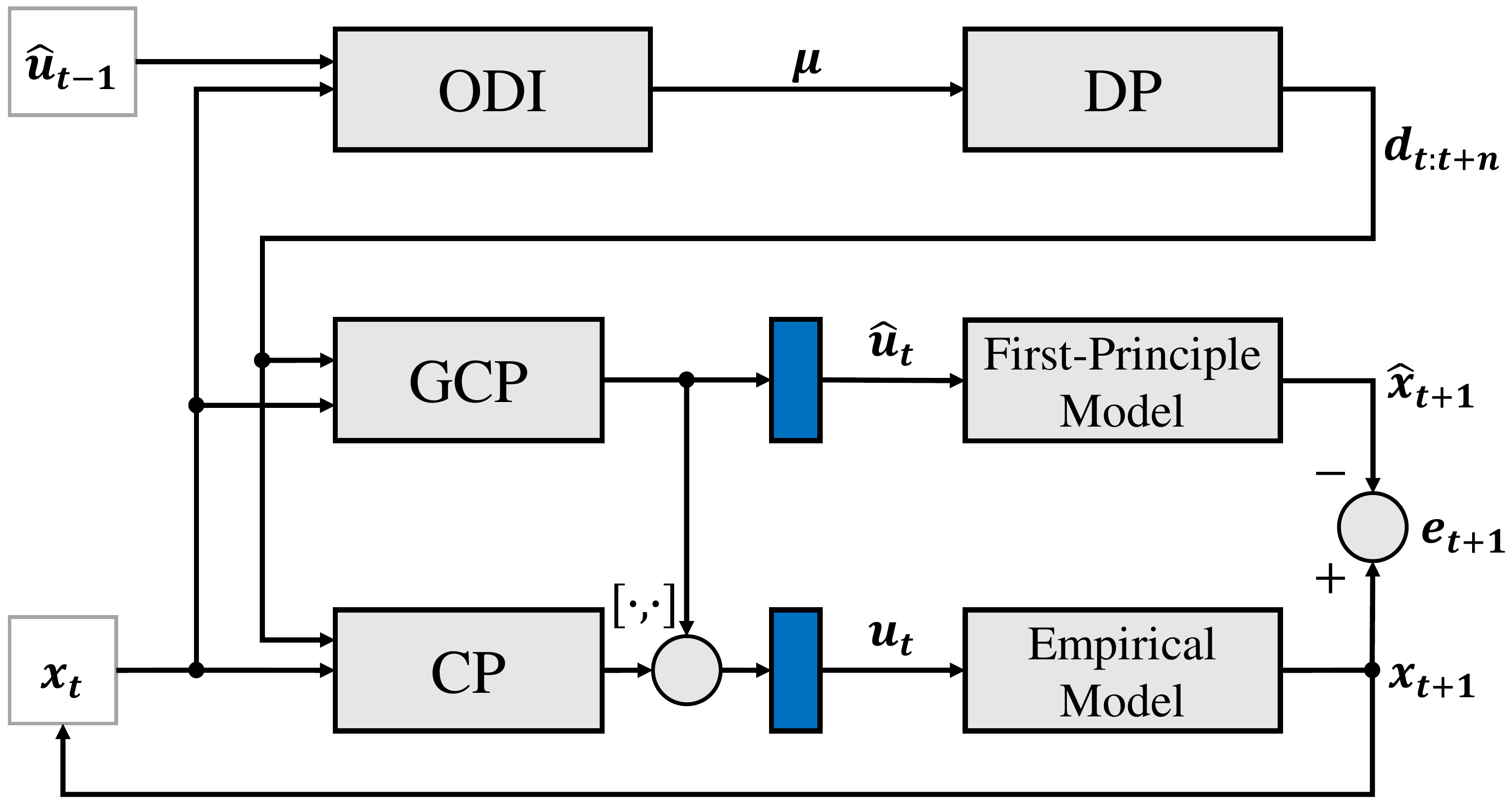}
	\caption{Diagram of transfer reinforcement learning using Transition Mismatch Compensation with Feature Combination (TMC-feature), $[\cdot,\cdot]$ represents feature combination, the blue blocks represent the final layers of two policy networks.}
	\label{fig_algorithm_diagram_tmc_feature}
\end{figure}

We then propose TMC with feature combination (see Fig.~\ref{fig_algorithm_diagram_tmc_feature}) to further improve TMC framework. Similar with TMC-control, the optimization of TMC-feature still applies the reward function \eqref{eqn_mismatch_reward} for maximizing task performance as well as minimizing transition mismatch at the same time.
The difference is that, rather than naively adding together the control actions, TMC-feature combines features before the last layer of the two policy networks, then feeds the combined features into fully connected layers to produce expected control. Combining middle layer features might be an effective way to deal with the control constraints, such approach can offer more flexibility and improve network capacity, since the middle layer features may contain more comprehensive information compared with the final layer outputs with physical significance. Through combining middle layer features, the restrict limitations in the control space might be eased in a "feature space", then reducing the learning difficulty of the compensatory policy.

\section{Experiments}
\label{section_experiments}


\subsection{Construction of Empirical Model}
\label{section_experiments_empirical}

\begin{figure}[b]
	\centering
	\includegraphics[width=\linewidth]{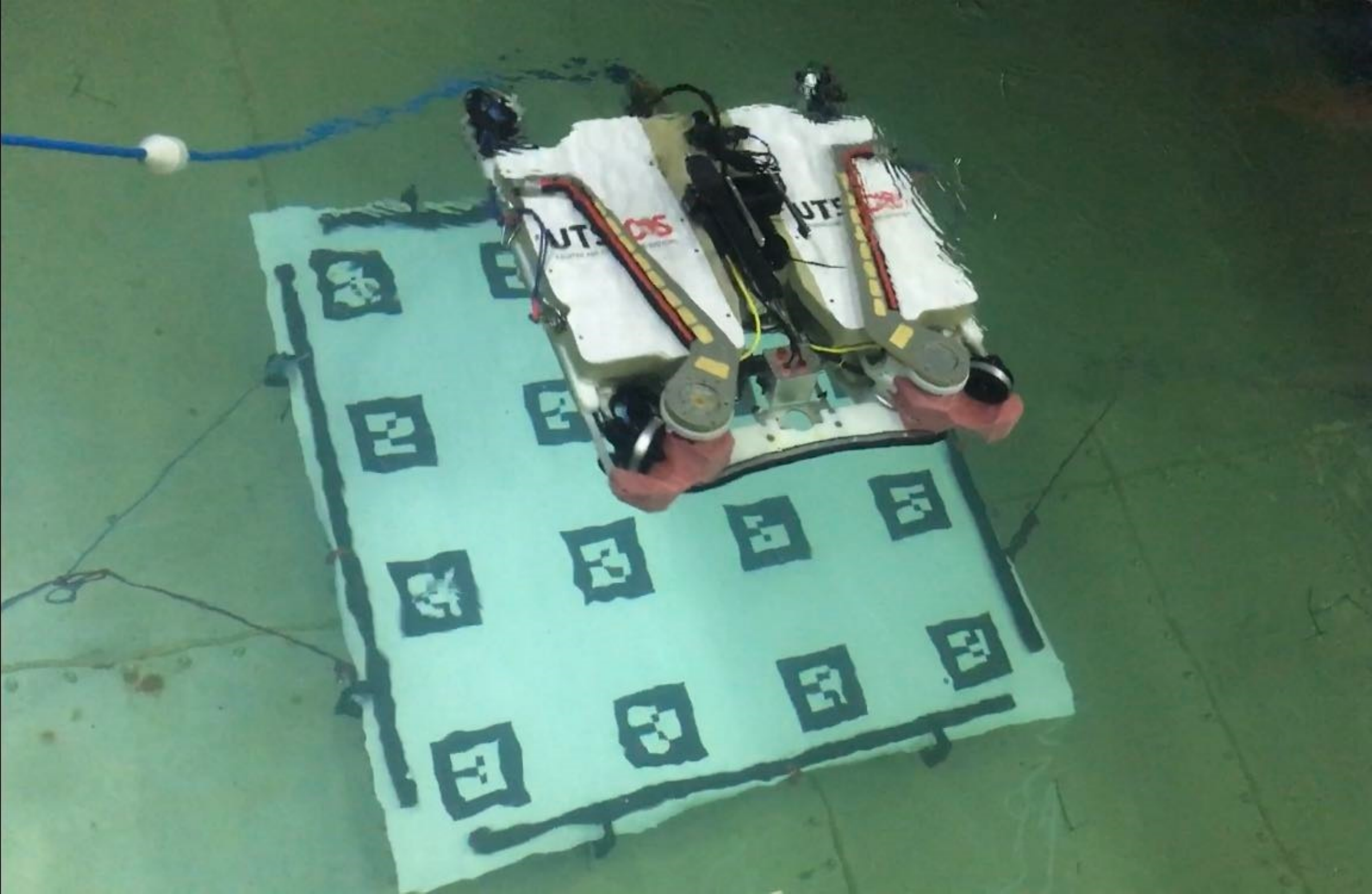}
	\caption{Tank test of Submerged Pile Inspection Robot (SPIR) over a set of QR code markers in a water tank.}
	\label{fig_field_test}
\end{figure}

\begin{table*}[t]
	\caption{Distributions of simulated disturbance parameters used in the source task.}
	\label{table_simulated_disturbance}
	\centering
	\begin{tabular}{|c|c|c|c|c|c|}
		\hline
		\textbf{Component Wave} & \textbf{1} & \textbf{2} & \textbf{3} & \textbf{4} & \textbf{5}\\
		\hline
		Amplitude w.r.t. $\left|\overline{\bm{u}}\right|$ & $0 \sim 50\%$ & $50 \sim 100\%$ & $50 \sim 100\%$ & $50 \sim 100\%$ & $0 \sim 50\%$\\
		\hline
		Period (s) & $1 \sim 2$ & $2 \sim 4$ & $2 \sim 4$ & $2 \sim 4$ & $4 \sim 8$\\
		\hline
		Phase (rad) & $-\pi \sim \pi$ & $-\pi \sim \pi$ & $-\pi \sim \pi$ & $-\pi \sim \pi$ & $-\pi \sim \pi$\\
		\hline
	\end{tabular}
\end{table*}

We are interested in deploying an underwater robot in shallow water where there are excessive wave forces. But it can be difficult to implement an accurate localization system in open water. Thus, we build a simulator based on an empirical model. Specifically, we collect motion data of an underwater robot, named Submerged Pile Inspection Robot (SPIR) (as shown in Fig.~\ref{fig_field_test}), in a water tank and train a deep neural network to represent the internal dynamics of the real robot (introduced in Section~\ref{section_experiments_empirical}). We also collect real-world wave forces in open water to represent the external disturbances (introduced in Section~\ref{section_experiments_setup}). The simulator is then used to define the target task.

We first build a visual localization system through using a downward-looking camera mounted to the underwater robot SPIR and a set of QR code markers fixed on the bottom of the water tank. The localization algorithm is able to calculate the relative pose of the robot with respect to a global frame defined by the markers. Fig.~\ref{fig_field_test} provides a demonstration of the tank test when SPIR localizes itself over the markers in the water tank.
Inertial Measurement Unit (IMU) data is also recorded.
SPIR is controlled by 12 thrusters, the control data of each thruster is directly read from the onboard computer in the form of Pulse-Width Modulation (PWM) signal.
During data collection, the robot moves in a fully autonomous mode and executes random control actions at each timestep. In order to ensure safe operation, an additional control term is employed when necessary to keep the robot away from the tank wall. We continuously record the motion and control data for over 4 hours.
The online data collection provides us with the raw sensor data expressed in their own frame. We then employ Extended Kalman Filter (EKF) \cite{zarchan2013fundamentals} to produce a more accurate estimation of the full state $\bm{x}$ of SPIR from these raw sensor data.

\begin{figure}[t]
	\centering
	\includegraphics[width=\linewidth]{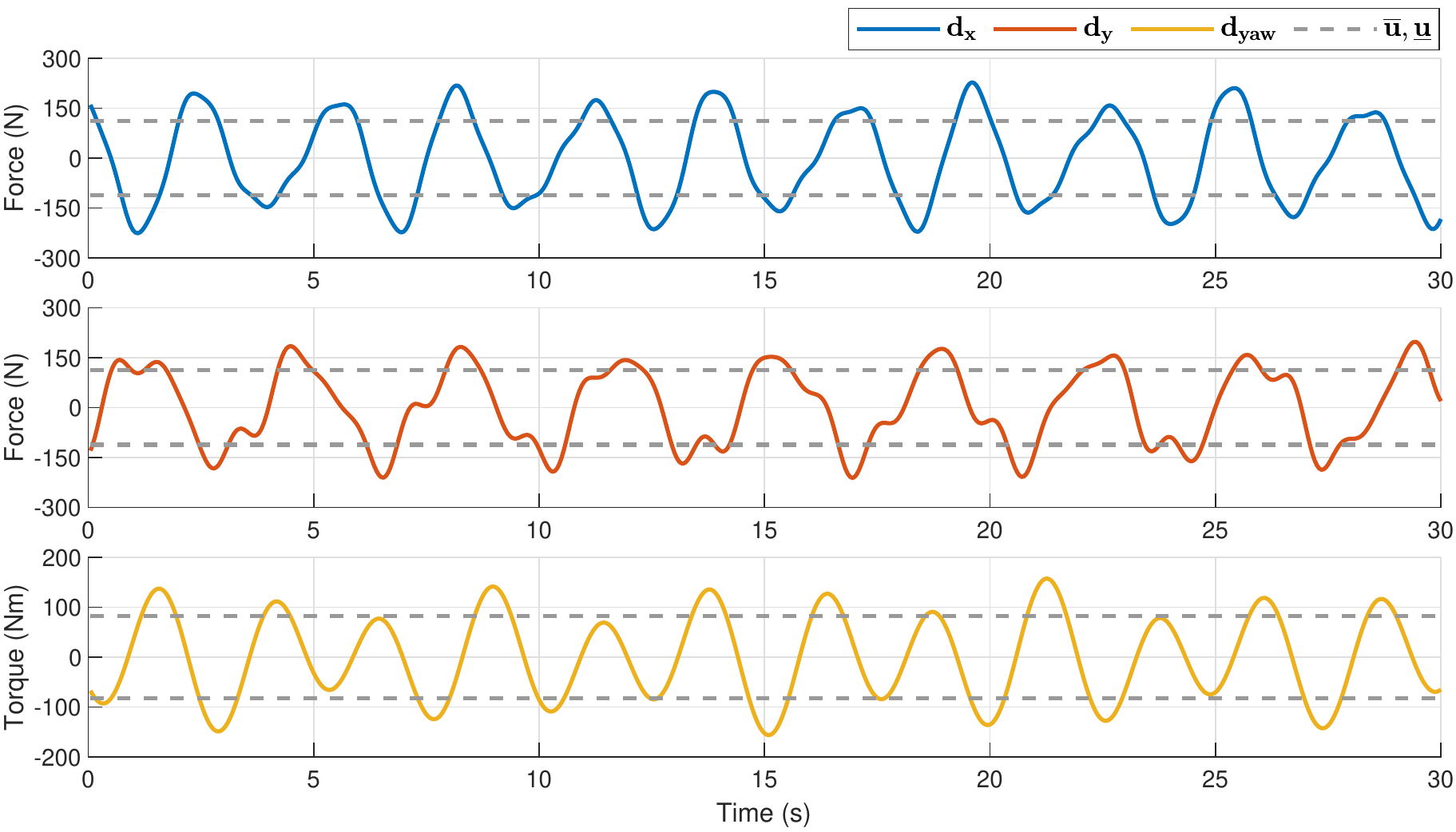}
	\caption{Example of simulated disturbances in X, Y and yaw directions.}
	\label{fig_simulated_disturbance}
\end{figure}

\begin{figure}[t]
	\centering
	\includegraphics[width=\linewidth]{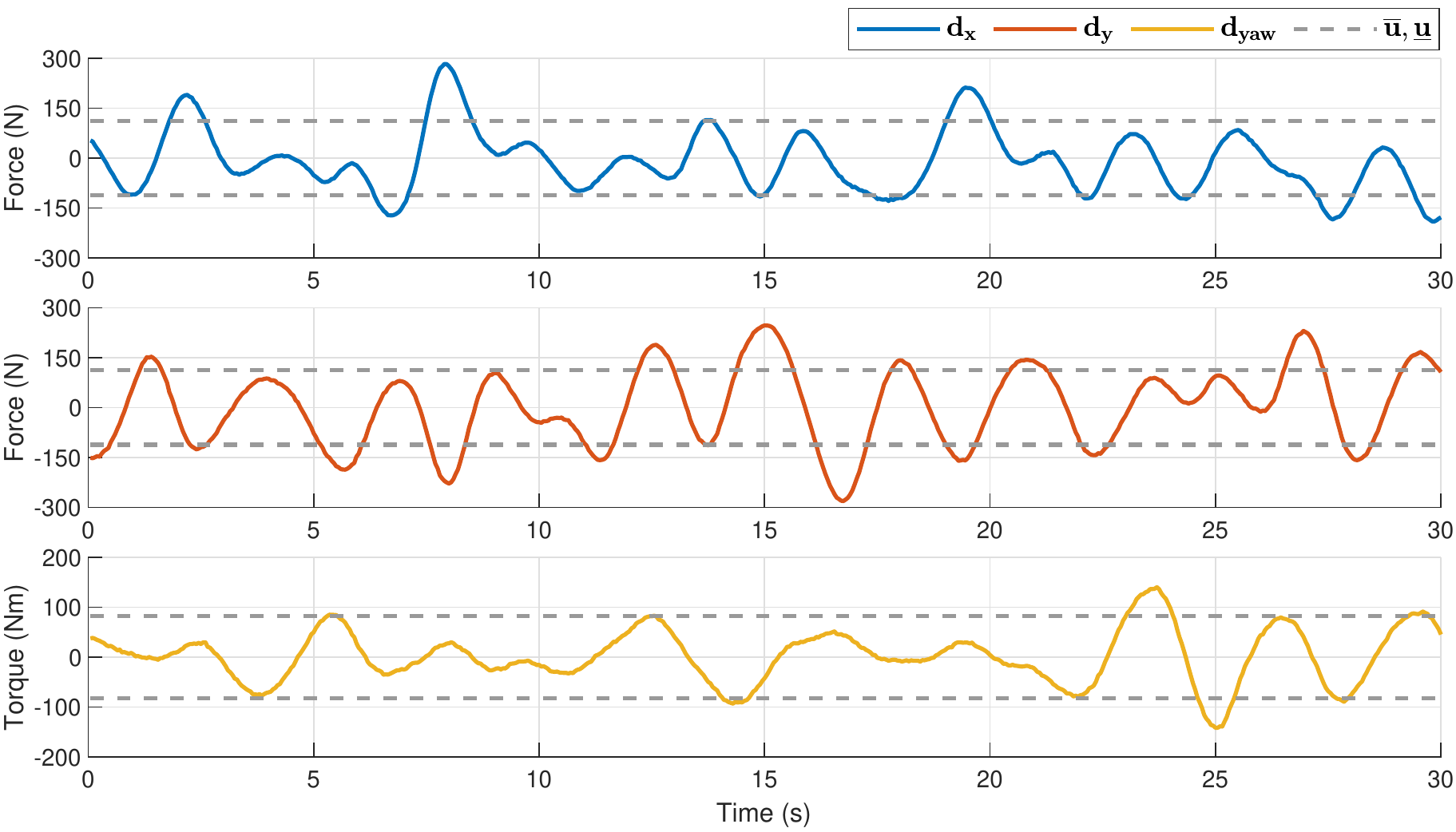}
	\caption{Example of real-world disturbances in X, Y and yaw directions.}
	\label{fig_real_disturbance}
\end{figure}

Normally, a transition function (i.e. dynamics model) would take the current state $\bm{x}_{t}$ and control $\bm{u}_{t}$ as input, and output the predicted next state $\bm{x}_{t+1}$. However, when there is a very short time interval $\Delta t$ between two consecutive states, the predicted next state $\bm{x}_{t+1}$ will become too similar with the current state $\bm{x}_{t}$, and the state difference may not well indicate the underlying dynamics \cite{nagabandi2018neural}. Thus, the transition function might be difficult to learn. This problem can be solved by learning a transition function that predicts the state change over one timestep $\Delta t$. Then the predicted next state is now given by: $\bm{x}_{t+1} = \bm{x}_{t} + f_{nn}\left(\bm{x}_{t},\bm{u}_{t};\theta_{f}\right)$.

We divided the recorded motion data into training dataset $\mathcal{D}_{train}$ and validation dataset $\mathcal{D}_{val}$, where the data is further sliced into inputs $\left(\bm{x}_{t},\bm{u}_{t}\right)$ and labels $\bm{x}_{t+1}-\bm{x}_{t}$. We then conduct feature normalization on both inputs and labels by subtracting the mean of the data and dividing by the standard deviation of the data, to ensure the loss function weights the different parts equally. 
After data preprocessing, we train the dynamics model $f_{nn}\left(\bm{x}_{t},\bm{u}_{t};\theta_{f}\right)$ offline using supervised learning by minimizing the error:
\begin{equation}
	\label{eqn_empirical_model_objective}
	\mathcal{E}(\theta_{f}) = \sum_{\left(\bm{x},\bm{u}\right) \in \mathcal{D}_{train}} \left\|\left(\bm{x}_{t+1}-\bm{x}_{t}\right)-f_{nn}\left(\bm{x}_{t},\bm{u}_{t};\theta_{f}\right)\right\|^{2},
\end{equation}
using SGD. While training on the training dataset $\mathcal{D}_{train}$, we also calculate the mean squared error in \eqref{eqn_empirical_model_objective} on the validation dataset $\mathcal{D}_{val}$ to optimize hyperparameters.

\subsection{Experimental Setup}
\label{section_experiments_setup}

The performance of the designed algorithms is tested through a pose regulation task in simulation, where the robot starts with random pose and velocity in each episode, the goal is to control the robot to navigate toward a target pose then stabilize itself under the external disturbances. Each episode contains 200 timesteps with $0.05s$ per timestep. The adopted A2C framework employs a parallel training mode through using 16 agents synchronously, the equivalent real-world training time for each agent in the source task is 16.67 hours.

Normally, an underwater robot has 6 degree of freedom (DOF). In this experiment, only the horizontal motion and control (X, Y and yaw) of the robot are considered, then the robot has a 6-dimensional state space $\mathcal{X} \in \mathcal{R}^{6}$ and a 3-dimensional action space $\mathcal{U} \in \mathcal{R}^{3}$.
In the source task, we build a first-principle model of SPIR that enables large scale training in simulation. The model is assumed to has the mass of $60kg$ and a simplified cuboid shape with the dimension of $0.68 \times 0.75 \times 0.19m^{3}$. Both hydrodynamics and control latency are excluded. The control constraints are given as $\left|\overline{\bm{u}}\right| = \left|\underline{\bm{u}}\right| = [112 N \ 112 N \ 82 Nm]^{T}$.

Besides, the source task uses simulated disturbances. They are exerted on all 3 directions (X, Y and yaw) of the robot in the global frame, and are constructed as a superposition of multiple sinusoidal waves with different amplitudes $A$, frequencies $\omega$ and phases $\phi$. In this work, we use a composition of 5 sinusoidal waves, whose parameter distributions and waveforms are given in Table~\ref{table_simulated_disturbance} and Fig.~\ref{fig_simulated_disturbance}. Our goal is to enable the trained control policy to deal with unknown time-correlated disturbances, and the policy is expected to adapt to a wide range of waveforms, instead of a fixed one. Thus, the amplitudes, frequencies and phases of the simulated disturbances are randomly sampled from the given distributions in each training episode.

The target task uses an empirical model of SPIR, composed of the neural network dynamics in Section~\ref{section_experiments_empirical} and real-world disturbances. These disturbances come from force data of real-world ocean waves collected in open water, as shown in Fig.~\ref{fig_real_disturbance}.
The data collection process is implemented through connecting a force/torque sensor between SPIR and a metal pole, the metal pole is fixed to a bridge over turbulent water flows, and the robot is deployed in the water and remains unactuated. Then, the readings of the force/torque sensor are regarded as the external wave forces and torques.
We notice that the real-world disturbances have widely varying amplitudes, which are not constrained within the ranges of the simulated disturbance parameters, leading to a more challenging control problem.

\subsection{Modular Network}

\begin{figure}[t]
	\centering
	\includegraphics[width=0.75\linewidth]{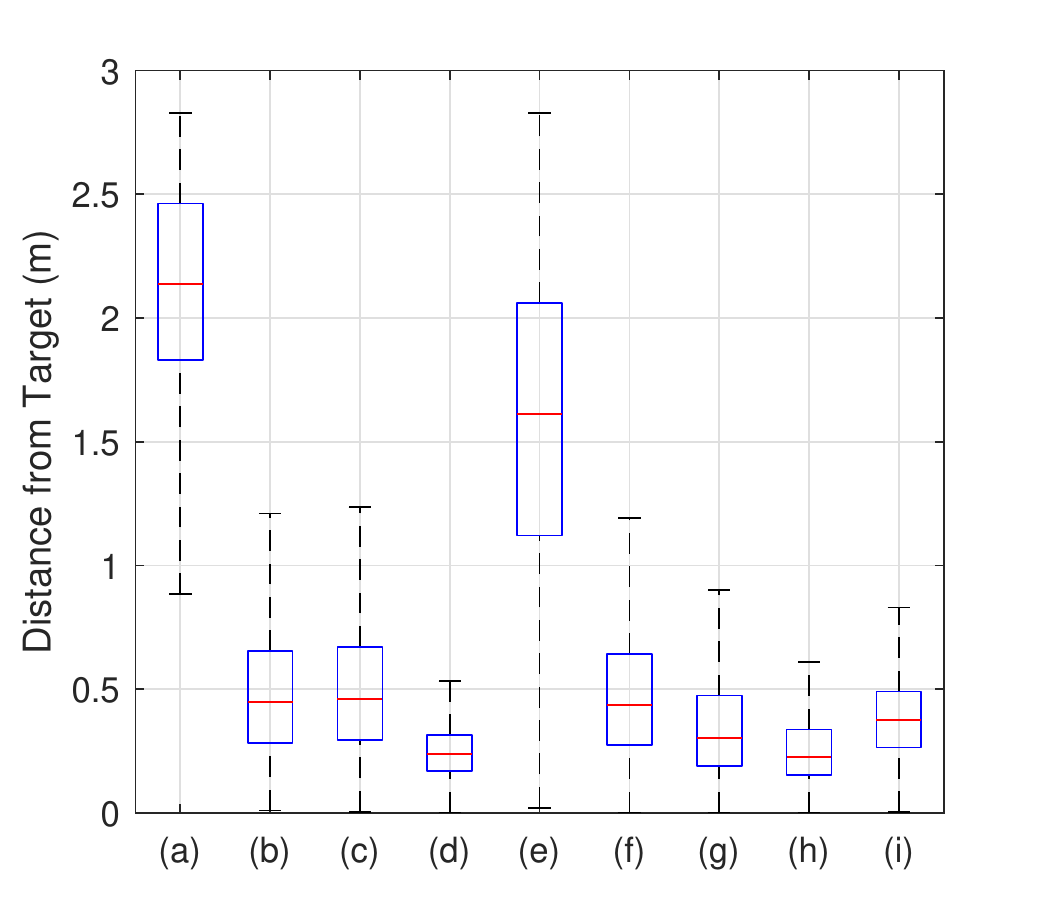}
	\caption{Distribution of the distance from the robot to the target during last 100 timesteps under the first-principle model: (a) A2C; (b) GCP-true using as input the frequency domain signals $\left\{A_{1},\omega_{1},\phi_{1},\cdots,A_{k},\omega_{k},\phi_{k}\right\}$; (c) GCP-true using as input the processed frequency domain signals $\left\{A_{1},\omega_{1},\omega_{1}t+\phi_{1},\cdots,A_{k},\omega_{k},\omega_{k}t+\phi_{k}\right\}$; (d) GCP-true using as input the time domain signals $\bm{d}_{t:t+n}$; (e) GCP-ODI at the 1st iteration; (f) GCP-ODI at the 2nd iteration; (g) GCP-ODI at the 3rd iteration; (h) GCP-ODI at the 4th iteration; (i) DOB-Net.}
	\label{fig_distance_distribution_gcp_odi}
\end{figure}

\begin{figure}[t]
	\centering
	\includegraphics[width=\linewidth]{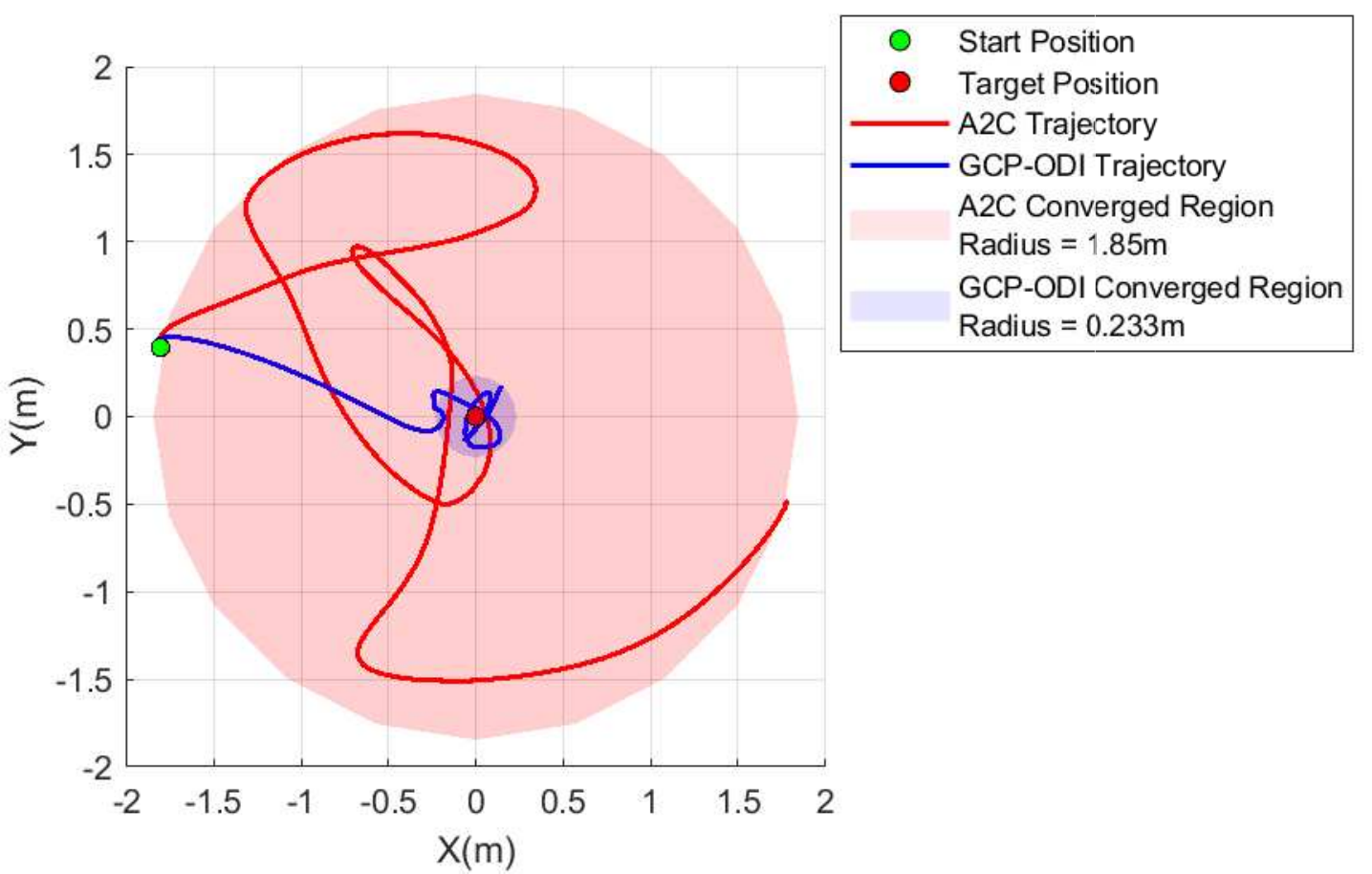}
	\caption{Trajectories of the robot using A2C and GCP-ODI under the first-principle model.}
	\label{fig_trajectory_gcp_odi}
\end{figure}

\begin{table*}[t]
	\caption{Variations in the first-principle model.}
	\label{table_model_variation}
	\centering
	\begin{tabular}{|c|c|c|}
		\hline
		\textbf{Parameters} & \textbf{Original Model} & \textbf{Variated Model}\\
		\hline
		\multirow{2}*{Mass} 
		& $60kg$ & $50kg$\\
		\cline{2-3}
		& $60kg$ & $70kg$\\
		\hline
		Geometry & \tabincell{c}{Cuboid Geometry: \\ $0.68 \times 0.75 \times 0.19 m^{3}$} & CAD Geometry\\
		\hline
		Hydrodynamics & No Added Mass or Damping & Added Mass and Damping\\
		\hline
		\multirow{2}*{\tabincell{c}{Velocity Constraints}} 
		& $\pm\left[1.0m/s \ 1.0m/s \ \frac{\pi}{2}rad/s\right]$ & $\pm\left[0.7m/s \ 0.7m/s \ \frac{\pi}{3}rad/s\right]$\\
		\cline{2-3}
		& $\pm\left[0.7m/s \ 0.7m/s \ \frac{\pi}{3}rad/s\right]$ & $\pm\left[1.0m/s \ 1.0m/s \ \frac{\pi}{2}rad/s\right]$\\
		\hline
		\multirow{2}*{\tabincell{c}{Control Constraints}} 
		& $\pm\left[112N \ 112N \ 82Nm\right]$ & $\pm\left[86N \ 86N \ 62Nm\right]$\\
		\cline{2-3}
		& $\pm\left[86N \ 86N \ 62Nm\right]$ & $\pm\left[112N \ 112N \ 82Nm\right]$\\
		\hline
		Control Offset & None & 30\% of Control Constraints $\left|\overline{\bm{u}}\right|$\\
		\hline
		Control Latency & None & $100ms$\\
		\hline
	\end{tabular}
\end{table*}

\begin{figure*}[t]
	\centering
	\subfloat[\scriptsize Smaller Mass]{
		\includegraphics[width=0.18\linewidth]{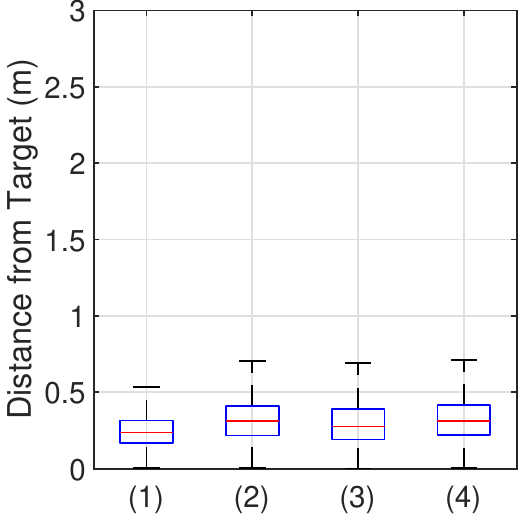}
	}
	\subfloat[\scriptsize Larger Mass]{
		\includegraphics[width=0.18\linewidth]{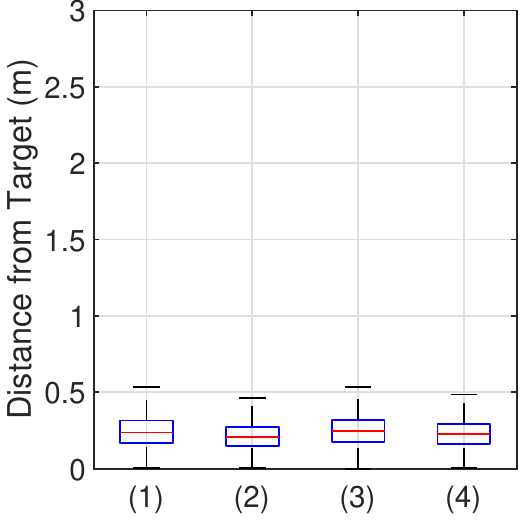}
	}
	\subfloat[\scriptsize Geometry]{
		\includegraphics[width=0.18\linewidth]{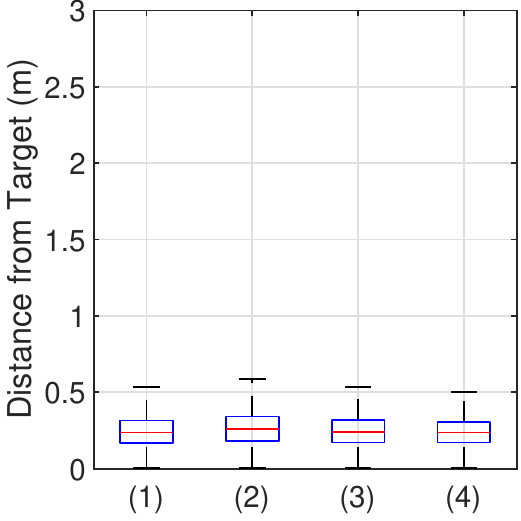}
	}
	\subfloat[\scriptsize Hydrodynamics]{
		\includegraphics[width=0.18\linewidth]{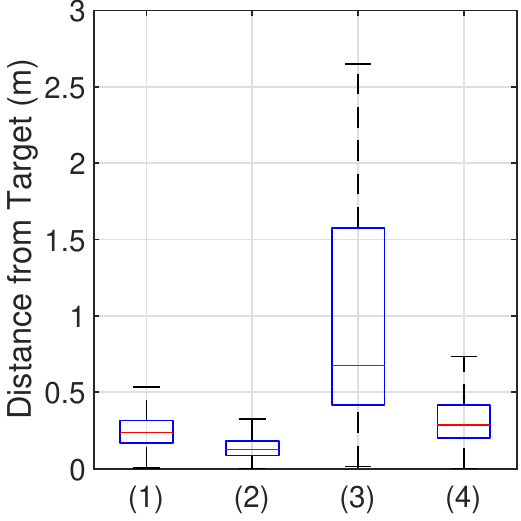}
	}
	\subfloat[\scriptsize Smaller Velocity Constraints]{
		\includegraphics[width=0.18\linewidth]{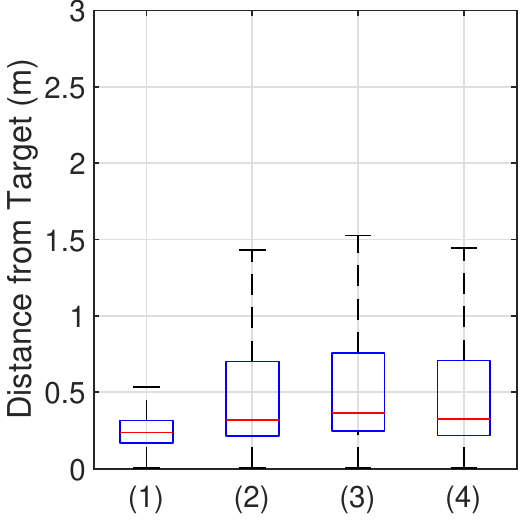}
	}
	\\
	\subfloat[\scriptsize Larger Velocity Constraints]{
		\includegraphics[width=0.18\linewidth]{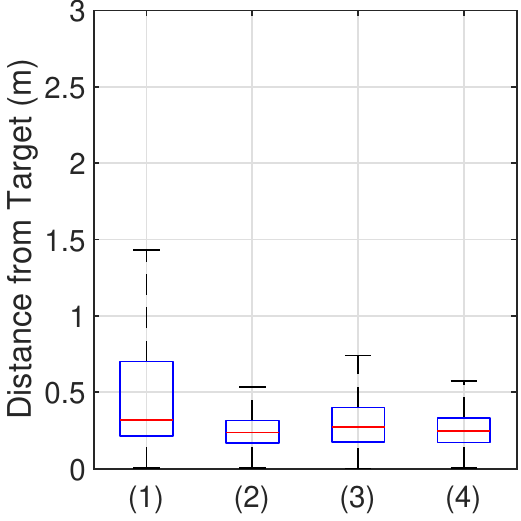}
	}
	\subfloat[\scriptsize Smaller Control Constraints]{
		\includegraphics[width=0.18\linewidth]{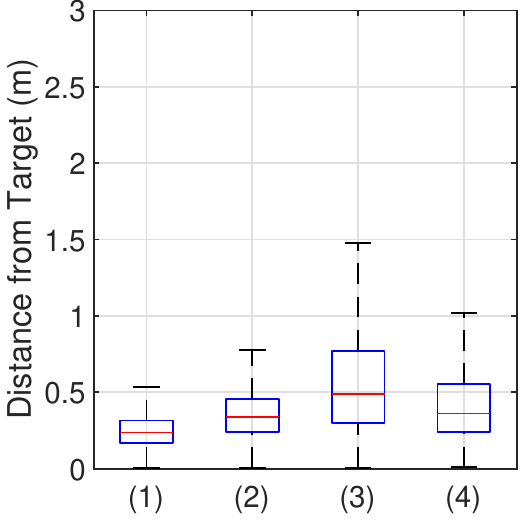}
	}
	\subfloat[\scriptsize Larger Control Constraints]{
		\includegraphics[width=0.18\linewidth]{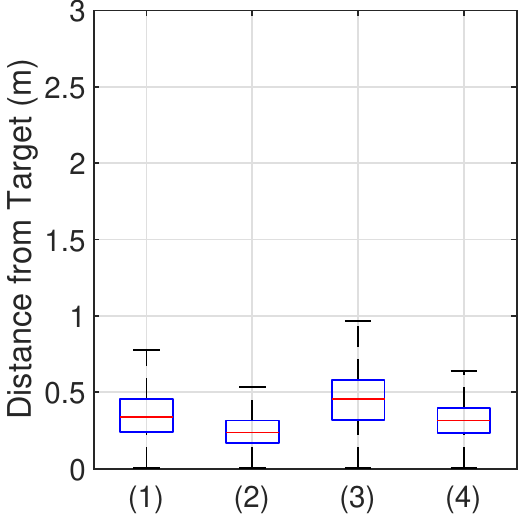}
	}
	\subfloat[\scriptsize Control Offset]{
		\includegraphics[width=0.18\linewidth]{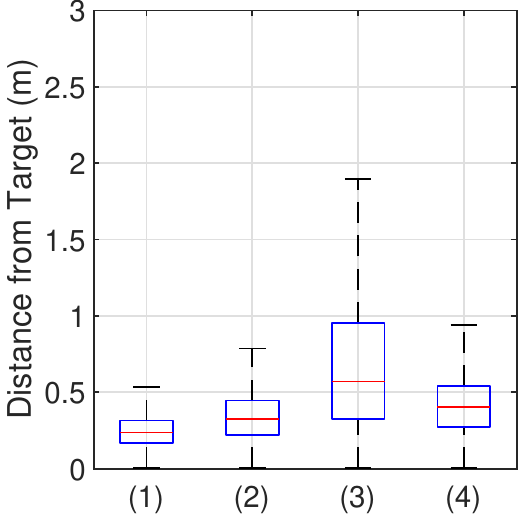}
	}
	\subfloat[\scriptsize Control Latency]{
		\includegraphics[width=0.18\linewidth]{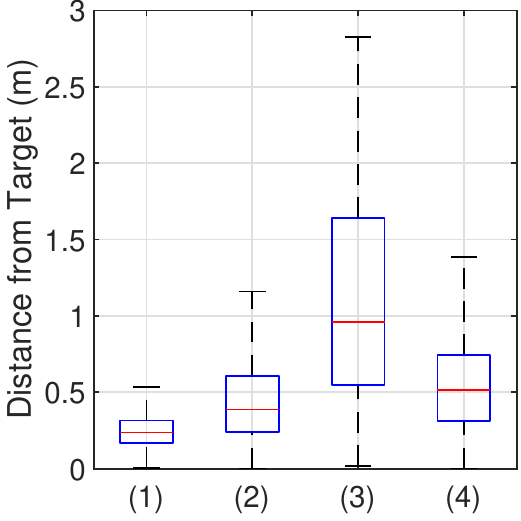}
	}
	\caption{Distribution of the distance from the robot to the target during last 100 timesteps for different variations in the first-principle model: (1) GCP-ODI trained and tested on the original model (i.e. source policy); (2) GCP-ODI trained and tested on the variated model (i.e. target policy); (3) GCP-ODI trained on the original model and tested on the variated model (i.e. unadapted policy); (4) transfer learning using TMC-feature (i.e. adapted policy).}
	\label{fig_distance_distribution_transfer_variation}
\end{figure*}

The performance of the modular learning architecture, GCP-ODI, is evaluated first for excessive disturbance rejection under the first-principle model.
As shown in Fig.~\ref{fig_distance_distribution_gcp_odi}, it can be seen that training a RL policy with disturbances $\bm{d}_{t:t+n}$ or disturbance parameters $\bm{\mu}$ as additional input can achieve better performance, compared with conventional RL policy (A2C).
In terms of the part of policy input representing disturbances, we found that using time domain signals outperforms using frequency domain signals, even though time is combined with the frequency domain parameters to give phase information $\bm{\mu} = \left\{A_{1},\omega_{1},\omega_{1}t+\phi_{1},\cdots,A_{k},\omega_{k},\omega_{k}t+\phi_{k}\right\}$. 
This result indicates that it might be difficult for RL to find an effective mapping between the frequency domain signals of the disturbances and the control output. In contrast, the time domain signals are more closely related to control. Thus, although the time domain variables only contain partial information of the disturbance waveforms (i.e. only $n$ timesteps of future disturbances), adopting these variables still perform well.

We also found that, the robot hardly stabilize itself near the target when ODI is only trained on the initially collected data (1st iteration). This is because these data is generated by a control policy and a dynamics model with consistent disturbance parameters $\bar{\bm{\mu}}$, but there are actually mismatched disturbance parameters $\bar{\bm{\mu}} \neq \hat{\bm{\mu}}$ during the online operation of GCP-ODI. This situation can be mitigated through iteratively alternating between gathering more data with the current ODI and GCP, and retraining ODI using the aggregated data. After several iterations (4 in our case), the disturbance rejection capability of the robot reaches a relatively high level, even approaching the performance of GCP when given the true disturbance parameters $\bar{\bm{\mu}}$.
In addition, we also evaluate an end-to-end learning framework for excessive disturbance rejection, called Disturbance Observer Network (DOB-Net) \cite{wang2019dob}, as a comparison with the modular architecture of GCP-ODI. We found that DOB-Net also achieves an excellent stability under the same task settings, but not as good as the iteratively trained GCP-ODI (4th iteration). This result is reasonable, because GCP explicitly takes the values of disturbance forces as input, instead of implicitly encoding the prediction of disturbances into the hidden state of RNN, as is the case in end-to-end learning. Such additional input information allows the policy to better specialize at different disturbance waveforms, thus potentially improves the control performance.

Fig.~\ref{fig_trajectory_gcp_odi} provides a more intuitive illustration of the algorithm effectiveness, by comparing the trajectories of the robot using the conventional A2C and the well trained GCP-ODI (after 4 iterations of training). We set a performance metric to be the range of the robot's distance to the target during last 100 timesteps of an episode, referred to as "converged region". It is found that GCP-ODI can dramatically improve the control stability of the robot under the unknown excessive disturbances, the converged region can be reduced from $1.85m$ to $0.233m$.

\subsection{Transfer Learning on Various Model Uncertainties}

It has been validated in the previous section that GCP-ODI performs well on the first-principle model,
but there are still many uncertainties in the actual system dynamics. Before deploying the transfer learning on the empirical model, we first evaluate the influence of different sources of model uncertainties.

In this part of evaluation, both the source and target tasks apply the first-principle models but with different model parameters, they are called original model and variated model, respectively. Possible uncertainties in the dynamics model for the underwater robot may include:
\begin{itemize}
	\item Mass
	
	\item Geometry
	
	\item Hydrodynamics
	
	\item Velocity Constraints
	
	\item Control Constraints
	
	\item Control Offset
	
	\item Control Latency
\end{itemize}
Table~\ref{table_model_variation} summarizes different model parameters and corresponding variations in the first-principle model. Fig.~\ref{fig_distance_distribution_transfer_variation} gives the detailed evaluation results of each type of model variation.
We can see that variations of mass and geometry have little influence on the control performance when using the unadapted policy compared with using the target policy on the variated model, then there is no need to spend much effort on estimating precise mass and geometry of the real system when designing a source task (a first-principle model) for transfer.
For the variations of velocity constraints, we found that the target policy, the unadapted policy, and the adapted policy on the variated model have similar performance. That is to say, the control performance depends mostly on the dynamics model itself, rather than the applied algorithms. Thus, it is not necessary to apply the transfer learning under this kind of model uncertainties.

It is possible to model most of the hydrodynamic effects, but still impossible to quantify online, due to the difficulty in estimating coefficients.
Thus we do not make any assumption of the hydrodynamics in the original model. This variation in the dynamics model brings a great impact on the control stability when using the unadapted policy.
Another important source of uncertainties is control latency, which widely exists in most of mechanical and electronic systems. But just like the hydrodynamics, the control latency is also hard to be well simulated, so we assume no latency in the original model. The nature of latency might complicate the motion data used in ODI, leading to adverse effects on the control performance for the unadapted policy.
The application of the transfer learning is necessary when these two model uncertainties exist, which is normally the case for the underwater robots.

The control constraints of the real system may have some scaling or offset compared to the mathematically modeled ones. This phenomenon can be caused by the uncertainties in thrusters' dynamics. Also, the robot may be subjected to some unexpected external forces, like the pulling force of the power cable for tethered AUV. The variation of the control constraints and the occurrence of the control offset have some influence for the unadapted policy, in which case the transfer learning is also required. But this influence is only moderate since most of the control signals are at the bound values $\overline{\bm{u}}$ and $\underline{\bm{u}}$ (see Fig.~\ref{fig_control_comparison_transfer_empirical}).

\subsection{Transfer Learning on Empirical Model}

\begin{figure}[t]
	\centering
	\includegraphics[width=0.75\linewidth]{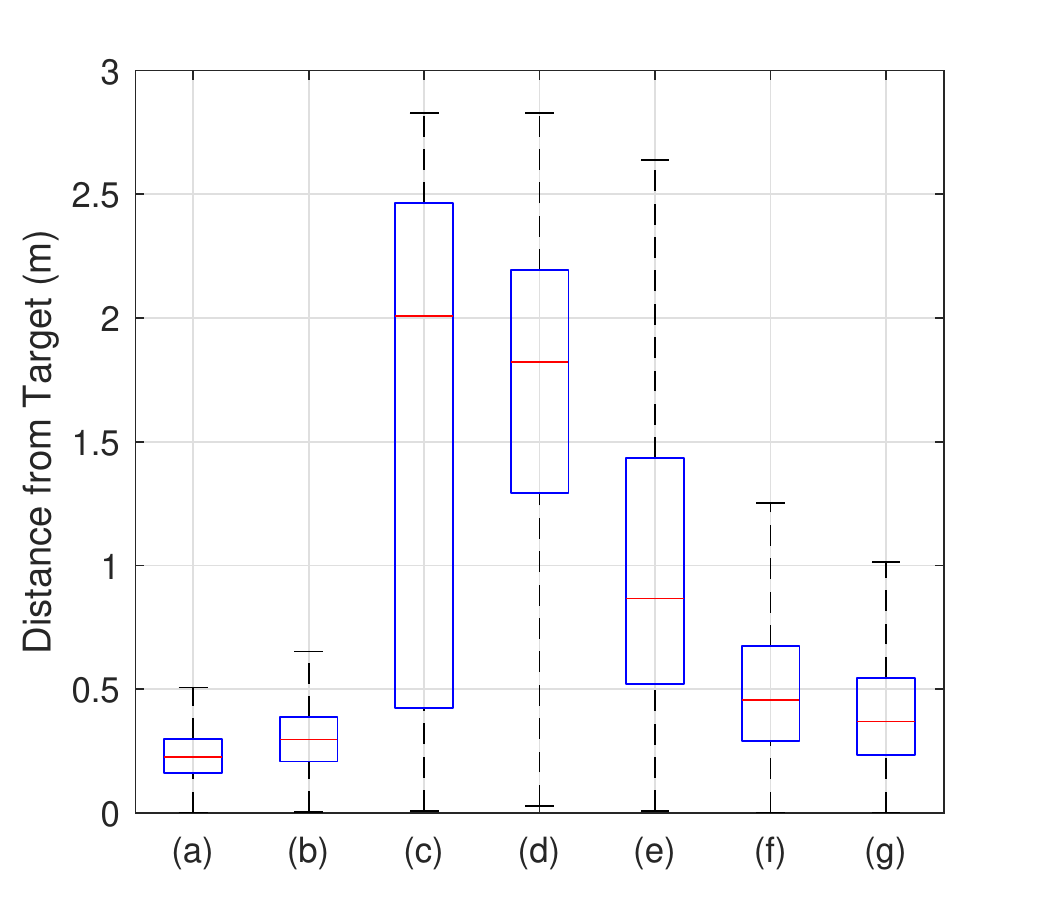}
	\caption{Distribution of the distance from the robot to the target during last 100 timesteps for the transfer learning: (a) GCP-ODI trained and tested on the first-principle model; (b) GCP-ODI trained and tested on the empirical model; (c) GCP-ODI trained on the first-principle model and tested on the empirical model; (d) transfer learning using CAC; (e) transfer learning using TMC-control; (f) transfer learning using TMC-feature; (g) transfer learning based on DOB-Net.}
	\label{fig_distance_distribution_transfer_empirical}
\end{figure}

\begin{figure}[t]
	\centering
	\includegraphics[width=\linewidth]{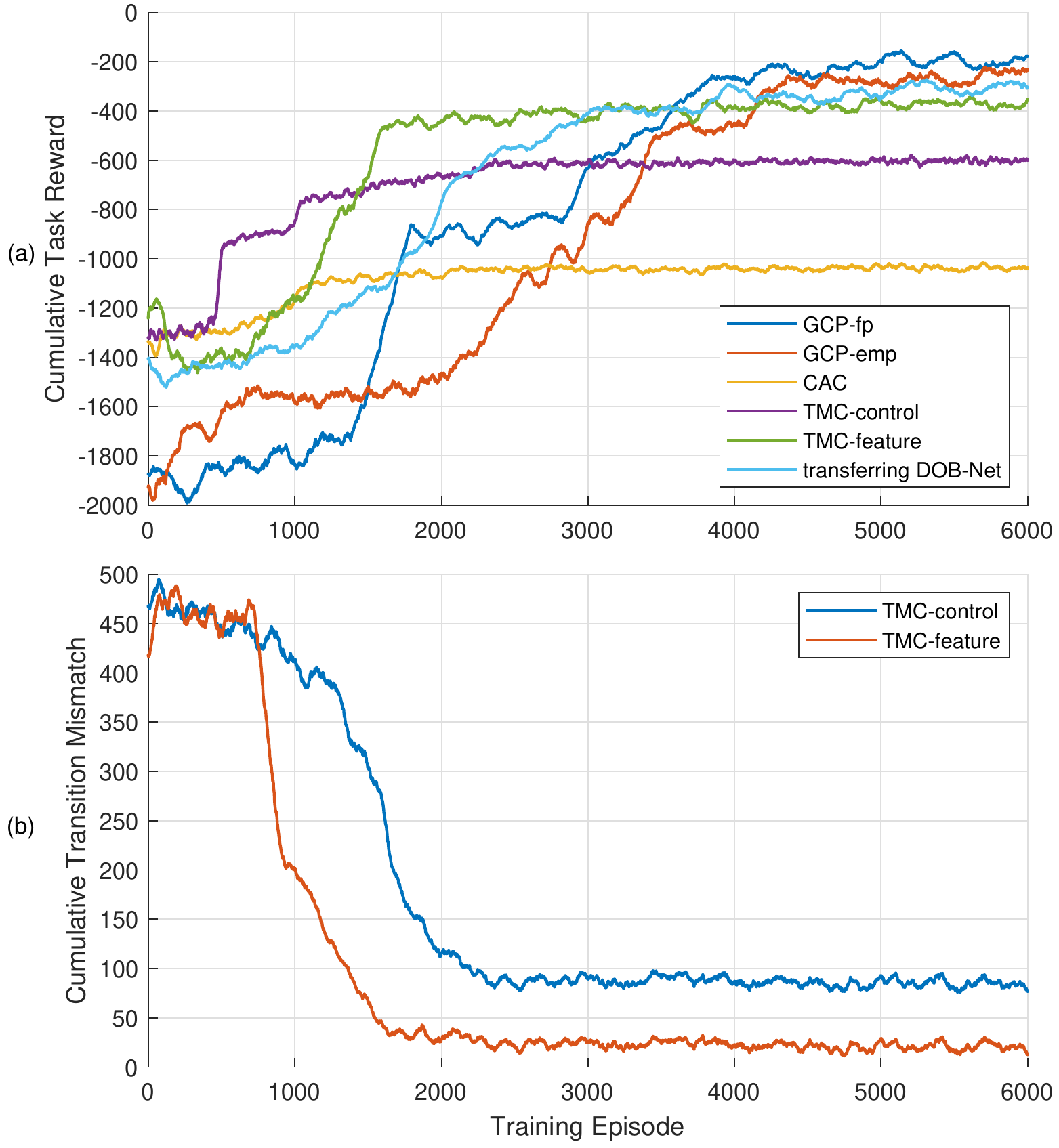}
	\caption{Training process of the transfer learning among different algorithms: (a) cumulative task reward; (b) cumulative transition mismatch.}
	\label{fig_training_reward_transfer_empirical}
\end{figure}

An empirical model has been constructed using real-world experimental data, 
then we focus on transferring a control policy trained on the first-principle model to be successfully deployed on the empirical model.
Note that we provide the results of GCP-ODI trained directly on the empirical model, these results are used as optimal performance in the target task, and can be considered as an informal performance upper bound for the transfer learning algorithms. This policy is available since the empirical model including the real-world disturbances can be deployed in the simulation, but these results are difficult to obtain on a real robot due to high sample complexity and damaging exploratory policy.

Fig.~\ref{fig_distance_distribution_transfer_empirical} shows the test results of different algorithms, we can see that GCP-ODI has poor stability when trained on the first-principle model then directly deployed on the empirical model. In contrast, GCP-ODI demonstrates much better performance when directly trained on the empirical model. That is the reason why we need the transfer learning between these two dynamics models.

The training process of the transfer learning can be visualized from Fig.~\ref{fig_training_reward_transfer_empirical}. We evaluate two objectives, which are cumulative task reward $\mathcal{R} = \sum_{t=1}^{T} r\left(\bm{x}_{t},\bm{u}_{t}\right)$ and cumulative transition mismatch $\mathcal{E} = \sum_{t=1}^{T} \left\|\bm{e}_{t}\right\|_{2}$ in each training episode.
It can be seen that the transfer learning algorithms converge faster than GCP using the true disturbance parameters $\bar{\bm{\mu}}$ on either the first-principle model or the empirical model (GCP-fp and GCP-emp), not to mention GCP is also followed by the iteratively training of ODI. Thus, the three algorithms of transfer learning based on GCP-ODI all improve the sample efficiency, compared with training GCP-ODI from scratch without transfer.

\begin{figure}[t]
	\centering
	\includegraphics[width=0.9\linewidth]{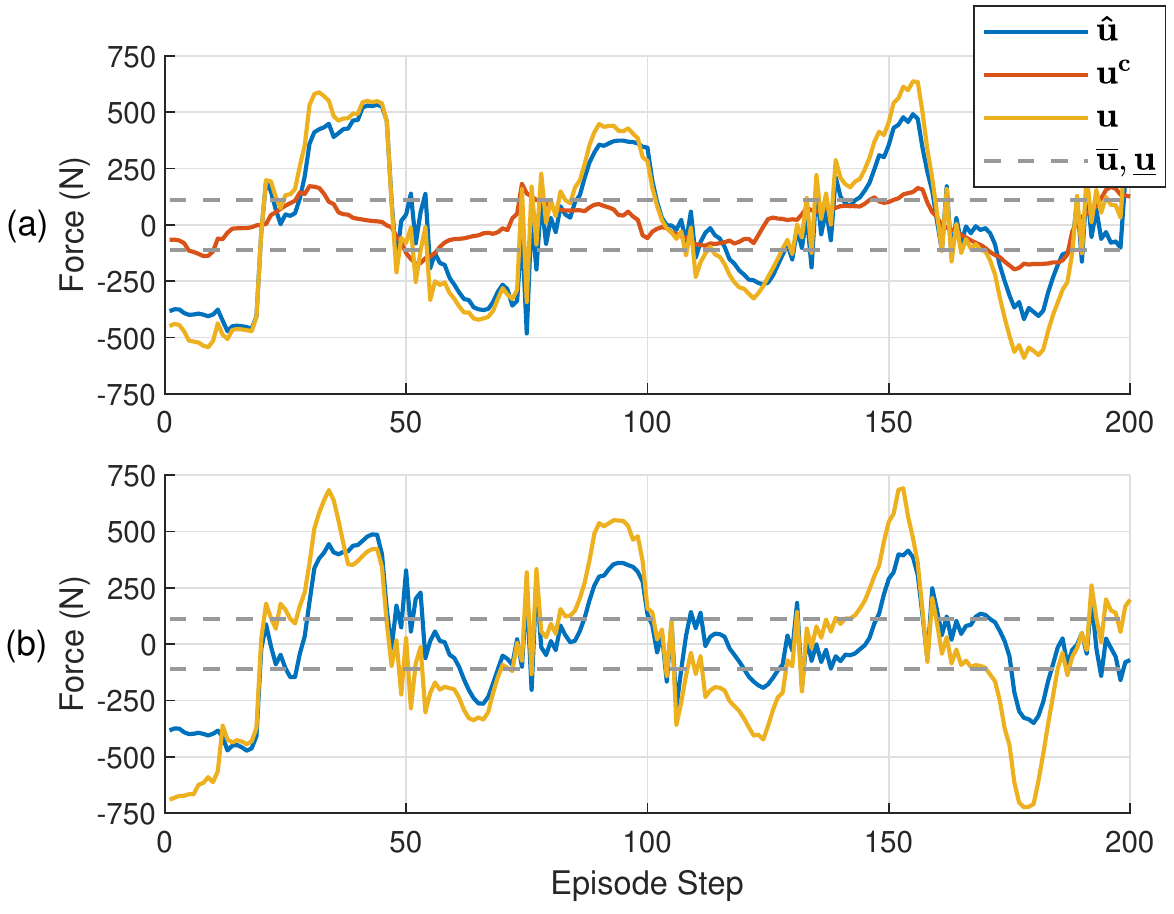}
	\caption{Comparison among different control signals (X-axis) for the robot during the transfer learning on the empirical model: (a) transfer learning using TMC-control; (b) transfer learning using TMC-feature.}
	\label{fig_control_comparison_transfer_empirical}
\end{figure}

\begin{figure}[t]
	\centering
	\includegraphics[width=\linewidth]{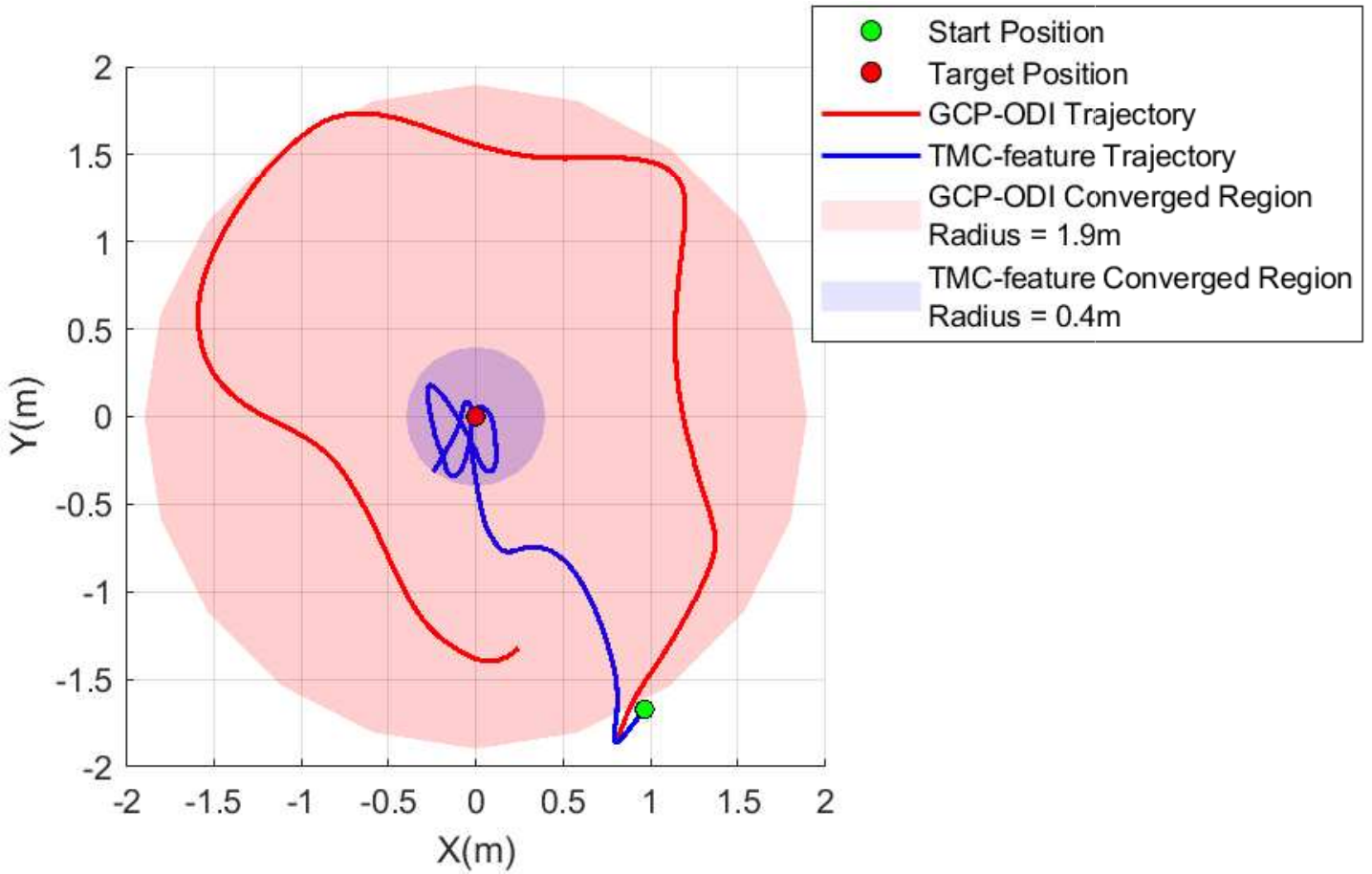}
	\caption{Trajectories of the robot using GCP-ODI and TMC-feature under the empirical model.}
	\label{fig_trajectory_transfer_empirical}
\end{figure}

Among the three transfer learning algorithms, CAC cannot successfully stabilize the robot, since ODI uses inconsistent trajectory data during transfer and training. 
TMC-control achieves much better performance due to the introduction of a parallel first-principle model, so that the transition mismatch $\bm{e}$ can be minimized. But this result still cannot be considered as a satisfactory solution.
We then look at TMC-feature, which gives a more effective combination approach of the compensatory control signals $\bm{u}^{c}$ and the original ones $\hat{\bm{u}}$, by combining middle layer features of the policy networks. Both Fig.~\ref{fig_training_reward_transfer_empirical} (a) and Fig.~\ref{fig_distance_distribution_transfer_empirical} prove that TMC-feature achieves better control performance than TMC-control.
In addition, through further evaluating the different control components during transfer (see Fig.~\ref{fig_control_comparison_transfer_empirical}), we found that in TMC-control algorithm, the compensation $\bm{u}^{c}$ is relatively small compared to GCP output $\hat{\bm{u}}$, then adding $\bm{u}^{c}$ to $\hat{\bm{u}}$ will not make much difference. Also, most of the added control actions have been truncated by the control constraints $\overline{\bm{u}}$ and $\underline{\bm{u}}$, leading to limited effects of the compensatory control actions. While for TMC-feature algorithm, combining features distinguishes the resultant control actions $\bm{u}$ from the output of GCP $\hat{\bm{u}}$, there is even obvious phase advance in $\bm{u}$ that compensates the control latency.
Fig.~\ref{fig_trajectory_transfer_empirical} shows the trajectories of the robot when using TMC-feature algorithm and the direct deployment of GCP-ODI on the empirical model after trained on the first-principle model. There is a significant improvement in the control stability of the robot after using the transfer learning, the converged region can be reduced to only $0.4m$.

However, even though the transfer learning using TMC-feature is able to well stabilize the robot, there is still noticeable gap from GCP-ODI directly trained on the empirical model (see Fig.~\ref{fig_distance_distribution_transfer_empirical}). This phenomenon can be explained through Fig.~\ref{fig_training_reward_transfer_empirical} (b), where the cumulative mismatch of TMC-feature cannot be fully eliminated. This result might be caused by the existence of control constraints, which limits the algorithm's ability to reject disturbances and compensate model mismatch in the meantime.

\begin{table*}[t]
	\caption{Predicted disturbance parameters by ODI under the empirical model.}
	\label{table_predicted_disturbance}
	\centering
	\begin{tabular}{|c|c|c|c|c|c|}
		\hline
		\textbf{Component Wave} & \textbf{1} & \textbf{2} & \textbf{3} & \textbf{4} & \textbf{5}\\
		\hline
		Amplitude w.r.t. $\left|\overline{\bm{u}}\right|$ & 34.06\%  & 66.63\% & 84.59\% & 53.93\% & 37.93\%\\
		\hline
		Period (s) & 5.98 & 3.01 & 2.73 & 2.31 & 1.88\\
		\hline
		Phase (rad) & -0.22$\pi$ & 0.40$\pi$ & -0.65$\pi$ & -0.94$\pi$ & -0.74$\pi$\\
		\hline
	\end{tabular}
\end{table*}

\begin{figure}[t]
	\centering
	\includegraphics[width=\linewidth]{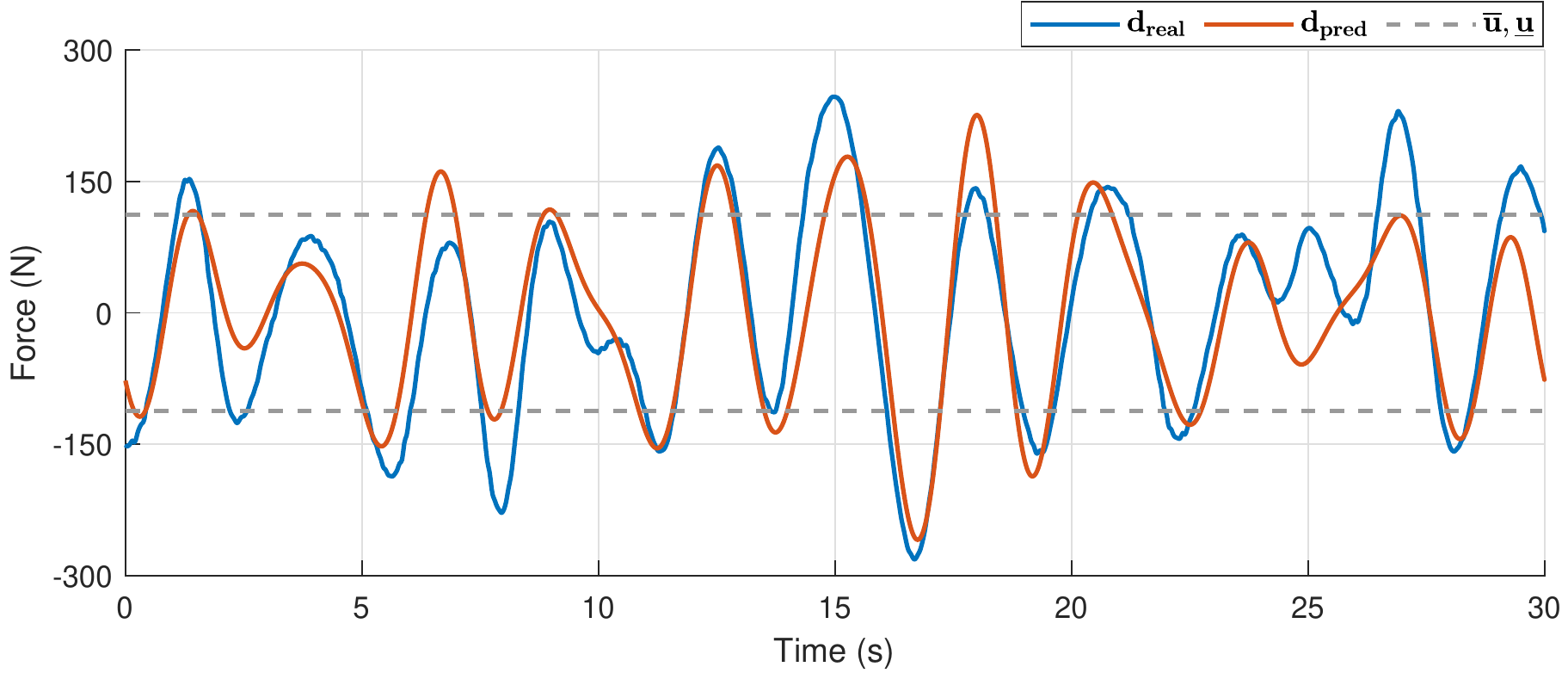}
	\caption{Comparison between the real-world disturbances and the predicted disturbances by ODI in Y direction, during transfer learning on the empirical model.}
	\label{fig_disturbance_comparison}
\end{figure}

In addition, during deployment on the empirical model, ODI is able to predict a set of disturbance parameters $\hat{\bm{\mu}}$ that best fit the real-world disturbances, which are not known to the learning algorithm and do not follow the distributions of simulated disturbance parameters. We give a waveform calculated from the disturbance parameters predicted at a timestep in the middle of a trajectory, as shown in Table~\ref{table_predicted_disturbance}, and compare it with the real-world disturbances exerted on the empirical model during test. We can see from Fig.~\ref{fig_disturbance_comparison} that the predicted disturbances are quite similar to the real-world ones, then there is only a small mismatch between these two disturbance waveforms. Thus, the total model mismatch that the transfer learning is required to compensate is reduced, then the burden for the compensation is reduced.

In this section, we also evaluate the transfer algorithm based on the end-to-end learning framework for excessive disturbance rejection, i.e. DOB-Net. Similar with TMC, this transfer process also trains an additional policy to compensate the dynamics model mismatch between the source and target tasks. This algorithm is used as a comparison with TMC.
Thanks to the ability of ODI to predict the unknown disturbances on the empirical model, the total model mismatch is reduced and the transfer learning is easier to train, compared with transfer learning based on DOB-Net. As illustrated in Fig.~\ref{fig_training_reward_transfer_empirical}, the convergence speed of TMC-feature is clearly higher than that of transferring DOB-Net. The final performance of TMC-feature, however, is not as good as the transferred DOB-Net (see Fig.~\ref{fig_distance_distribution_transfer_empirical}), this is reasonable since the transition mismatch $\bm{e}$ is difficult to be minimized to zero under current algorithm design. Further improvements could be investigated on a more optimized approach to combine compensatory signals.

\section{Conclusion \& Future Work}
\label{section_conclusion}

This paper proposes a learning framework to address the control problem for excessive disturbance rejection of an underwater robot under dynamics model mismatch. 
We first introduce a modular architecture of RL, composed of an observer network (ODI) and a controller network (GCP). ODI is used to predict a set of disturbance waveforms based on the observed past states and actions of the system, GCP then is able to produce expected control to actively reject disturbances from the current system state and the predicted disturbance waveforms.
Then we develop a transfer RL algorithm, TMC, based on the modular architecture that learns an additional compensatory policy through minimizing mismatch of transitions predicted by the two dynamics models of the source and target tasks. And the compensatory policy is added in terms of middle layer features instead of final network outputs in the target task.

The transfer learning algorithms are evaluated on a pose regulation task in simulation, where the source task defines a first-principle model of SPIR developed from the fundamental principles of dynamics, the target task applies an empirical model of SPIR derived from real-world experimental data. 
As a result, TMC algorithm achieves satisfactory performance on the empirical model after transfer, in the meantime enhance the sample efficiency with respect to learning from scratch without transfer. Furthermore, the modular architecture also outperforms the end-to-end network during transfer in terms of the sample efficiency, since the observer network ODI is able to predict the disturbances in the target task, then the total model mismatch is reduced.

Several aspects can be further explored in the future. 
It is known that if there are more component sinusoidal waves in simulated disturbances, then more realistic disturbance waveforms ODI can predict. But as the number of component waves increases (over 5), identifying high-dimensional disturbance parameters becomes challenging. Because it typically requires millions of training samples to cover a large output space of ODI, both trajectory data gathering by GCP and training of ODI have exponentially increased difficulty. More rigorous analysis is needed to evaluate the sample efficiency of GCP-ODI regarding high-dimensional disturbance parameters.

For both the disturbance rejection control using GCP-ODI and the tranfer learning using TMC, we can see that these policies can achieve similar performance with the corresponding baselines, which are GCP given the true disturbance parameters and GCP-ODI directly trained on the target task, respectively. However, the conditions of convergence and the theoretical limits of the proposed algorithms have not been established in this research, thus definitely require further investigation.

In the formulation of modular network architecture, GCP takes a fixed length of time domain disturbances as additional input, constructed from the predicted disturbance parameters of ODI. Such formulation outperforms using frequency domain signals (i.e. predicted disturbance parameters), but it provides only partial information of the disturbance waveforms, thus cannot ensure the control solutions are optimal. A more efficient representation of the disturbance information is required.
In addition, the real-world wave forces may be jointly determined by fluid conditions, robot morphology, as well as varying robot states and controls, and may vary with not only time but also space. Thus, using a superposition of multiple sinusoidal waves, which are only functions of time, to simulate the disturbances may not be sufficient. A comprehensive and multivariable function is required for a better description of the wave forces. We can seek a machine learning approach to explicitly build a wave model for interested water areas.

This research only covers the work using an empirical model in simulation, while the ultimate goal is to deploy the transfer learning on real-world robotic systems.
%
A possible idea is to use professional devices, like wave generators, to generate realistic wave forces in the water tank.
Another way is to directly deploy the robot in open water, where an accurate localization system would be necessary. Current solutions mainly focus on Simultaneous Localization And Mapping (SLAM), and the biggest concerns will be the localization accuracy on an unstable platform and the blurred images underwater.

The sample efficiency of the transfer learning directly depends on the mismatch between the first-principle model and the empirical model or the real robot. Sometimes, we may not be given a good prior model of the robot, then there is no way to pretrain a policy that helps in the transfer. In that case, we may investigate model-based reinforcement learning that online learns a model then optimizes a policy based on the model.


%



%
%

\ifCLASSOPTIONcaptionsoff
  \newpage
\fi

\IEEEtriggeratref{62}


\bibliographystyle{./bibliographies/IEEEtran}
\bibliography{./bibliographies/IEEEabrv,./bibliographies/mybibfile}

\begin{thebibliography}{10}
\providecommand{\url}[1]{#1}
\csname url@rmstyle\endcsname
\providecommand{\newblock}{\relax}
\providecommand{\bibinfo}[2]{#2}
\providecommand\BIBentrySTDinterwordspacing{\spaceskip=0pt\relax}
\providecommand\BIBentryALTinterwordstretchfactor{4}
\providecommand\BIBentryALTinterwordspacing{\spaceskip=\fontdimen2\font plus
\BIBentryALTinterwordstretchfactor\fontdimen3\font minus
  \fontdimen4\font\relax}
\providecommand\BIBforeignlanguage[2]{{%
\expandafter\ifx\csname l@#1\endcsname\relax
\typeout{** WARNING: IEEEtran.bst: No hyphenation pattern has been}%
\typeout{** loaded for the language `#1'. Using the pattern for}%
\typeout{** the default language instead.}%
\else
\language=\csname l@#1\endcsname
\fi
#2}}

\bibitem{antonelli2018underwater}
G.~Antonelli, \emph{Underwater robots}.\hskip 1em plus 0.5em minus 0.4em\relax
  Springer, 2018, vol. 123.

\bibitem{griffiths2002technology}
G.~Griffiths, \emph{Technology and applications of autonomous underwater
  vehicles}.\hskip 1em plus 0.5em minus 0.4em\relax CRC Press, 2002, vol.~2.

\bibitem{dean1991water}
R.~Dean and R.~Dalrymple, ``Water wave mechanics for scientists and
  engineers,'' \emph{Advanced Series on Ocean Engineering, World Scientific},
  vol.~2, 1991.

\bibitem{fernandez2016model}
D.~C. Fern{\'a}ndez and G.~A. Hollinger, ``Model predictive control for
  underwater robots in ocean waves,'' \emph{IEEE Robotics and Automation
  letters}, vol.~2, no.~1, pp. 88--95, 2016.

\bibitem{woolfrey2016kinematic}
J.~Woolfrey, D.~Liu, and M.~Carmichael, ``Kinematic control of an autonomous
  underwater vehicle-manipulator system (auvms) using autoregressive prediction
  of vehicle motion and model predictive control,'' in \emph{Robotics and
  Automation (ICRA), 2016 IEEE International Conference on}.\hskip 1em plus
  0.5em minus 0.4em\relax IEEE, 2016, pp. 4591--4596.

\bibitem{wang2018case}
T.~Wang, W.~Lu, and D.~Liu, ``A case study: Modeling of a passive flexible link
  on a floating platform for intervention tasks,'' in \emph{2018 13th World
  Congress on Intelligent Control and Automation (WCICA)}.\hskip 1em plus 0.5em
  minus 0.4em\relax IEEE, 2018, pp. 187--193.

\bibitem{xie2000much}
L.-L. Xie and L.~Guo, ``How much uncertainty can be dealt with by feedback?''
  \emph{IEEE Transactions on Automatic Control}, vol.~45, no.~12, pp.
  2203--2217, 2000.

\bibitem{gao2014centrality}
Z.~Gao, ``On the centrality of disturbance rejection in automatic control,''
  \emph{ISA transactions}, vol.~53, no.~4, pp. 850--857, 2014.

\bibitem{li2014disturbance}
S.~Li, J.~Yang, W.-H. Chen, and X.~Chen, \emph{Disturbance observer-based
  control: methods and applications}.\hskip 1em plus 0.5em minus 0.4em\relax
  CRC press, 2014.

\bibitem{chen2000nonlinear}
W.-H. Chen, D.~J. Ballance, P.~J. Gawthrop, and J.~O'Reilly, ``A nonlinear
  disturbance observer for robotic manipulators,'' \emph{IEEE Transactions on
  industrial Electronics}, vol.~47, no.~4, pp. 932--938, 2000.

\bibitem{chen2016disturbance}
W.-H. Chen, J.~Yang, L.~Guo, and S.~Li, ``Disturbance-observer-based control
  and related methods—an overview,'' \emph{IEEE Transactions on Industrial
  Electronics}, vol.~63, no.~2, pp. 1083--1095, 2016.

\bibitem{sun2016neural}
H.~Sun and L.~Guo, ``Neural network-based dobc for a class of nonlinear systems
  with unmatched disturbances,'' \emph{IEEE transactions on neural networks and
  learning systems}, vol.~28, no.~2, pp. 482--489, 2016.

\bibitem{garcia1989model}
C.~E. Garcia, D.~M. Prett, and M.~Morari, ``Model predictive control: theory
  and practice—a survey,'' \emph{Automatica}, vol.~25, no.~3, pp. 335--348,
  1989.

\bibitem{camacho2013model}
E.~F. Camacho and C.~B. Alba, \emph{Model predictive control}.\hskip 1em plus
  0.5em minus 0.4em\relax Springer Science \& Business Media, 2013.

\bibitem{gao2016nonlinear}
H.~Gao and Y.~Cai, ``Nonlinear disturbance observer-based model predictive
  control for a generic hypersonic vehicle,'' \emph{Proceedings of the
  Institution of Mechanical Engineers, Part I: Journal of Systems and Control
  Engineering}, vol. 230, no.~1, pp. 3--12, 2016.

\bibitem{sutton2018reinforcement}
R.~S. Sutton and A.~G. Barto, \emph{Reinforcement learning: An
  introduction}.\hskip 1em plus 0.5em minus 0.4em\relax MIT press, 2018.

\bibitem{mnih2015human}
V.~Mnih, K.~Kavukcuoglu, D.~Silver, A.~A. Rusu, J.~Veness, M.~G. Bellemare,
  A.~Graves, M.~Riedmiller, A.~K. Fidjeland, G.~Ostrovski, \emph{et~al.},
  ``Human-level control through deep reinforcement learning,'' \emph{Nature},
  vol. 518, no. 7540, p. 529, 2015.

\bibitem{oh2016control}
J.~Oh, V.~Chockalingam, S.~Singh, and H.~Lee, ``Control of memory, active
  perception, and action in minecraft,'' \emph{arXiv preprint
  arXiv:1605.09128}, 2016.

\bibitem{gu2016continuous}
S.~Gu, T.~Lillicrap, I.~Sutskever, and S.~Levine, ``Continuous deep q-learning
  with model-based acceleration,'' in \emph{International Conference on Machine
  Learning}, 2016, pp. 2829--2838.

\bibitem{schulman2015trust}
J.~Schulman, S.~Levine, P.~Abbeel, M.~Jordan, and P.~Moritz, ``Trust region
  policy optimization,'' in \emph{International Conference on Machine
  Learning}, 2015, pp. 1889--1897.

\bibitem{gu2016q}
S.~Gu, T.~Lillicrap, Z.~Ghahramani, R.~E. Turner, and S.~Levine, ``Q-prop:
  Sample-efficient policy gradient with an off-policy critic,'' \emph{arXiv
  preprint arXiv:1611.02247}, 2016.

\bibitem{lillicrap2015continuous}
T.~P. Lillicrap, J.~J. Hunt, A.~Pritzel, N.~Heess, T.~Erez, Y.~Tassa,
  D.~Silver, and D.~Wierstra, ``Continuous control with deep reinforcement
  learning,'' \emph{arXiv preprint arXiv:1509.02971}, 2015.

\bibitem{mnih2016asynchronous}
V.~Mnih, A.~P. Badia, M.~Mirza, A.~Graves, T.~Lillicrap, T.~Harley, D.~Silver,
  and K.~Kavukcuoglu, ``Asynchronous methods for deep reinforcement learning,''
  in \emph{International conference on machine learning}, 2016, pp. 1928--1937.

\bibitem{schulman2015high}
J.~Schulman, P.~Moritz, S.~Levine, M.~Jordan, and P.~Abbeel, ``High-dimensional
  continuous control using generalized advantage estimation,'' \emph{arXiv
  preprint arXiv:1506.02438}, 2015.

\bibitem{kaelbling1998planning}
L.~P. Kaelbling, M.~L. Littman, and A.~R. Cassandra, ``Planning and acting in
  partially observable stochastic domains,'' \emph{Artificial intelligence},
  vol. 101, no. 1-2, pp. 99--134, 1998.

\bibitem{shani2013survey}
G.~Shani, J.~Pineau, and R.~Kaplow, ``A survey of point-based pomdp solvers,''
  \emph{Autonomous Agents and Multi-Agent Systems}, vol.~27, no.~1, pp. 1--51,
  2013.

\bibitem{wierstra2010recurrent}
D.~Wierstra, A.~F{\"o}rster, J.~Peters, and J.~Schmidhuber, ``Recurrent policy
  gradients,'' \emph{Logic Journal of the IGPL}, vol.~18, no.~5, pp. 620--634,
  2010.

\bibitem{hausknecht2015deep}
M.~Hausknecht and P.~Stone, ``Deep recurrent q-learning for partially
  observable mdps,'' in \emph{2015 AAAI Fall Symposium Series}, 2015.

\bibitem{heess2015memory}
N.~Heess, J.~J. Hunt, T.~P. Lillicrap, and D.~Silver, ``Memory-based control
  with recurrent neural networks,'' \emph{arXiv preprint arXiv:1512.04455},
  2015.

\bibitem{sorokin2015deep}
I.~Sorokin, A.~Seleznev, M.~Pavlov, A.~Fedorov, and A.~Ignateva, ``Deep
  attention recurrent q-network,'' \emph{arXiv preprint arXiv:1512.01693},
  2015.

\bibitem{wang2018excessive}
T.~Wang, W.~Lu, and D.~Liu, ``Excessive disturbance rejection control of
  autonomous underwater vehicle using reinforcement learning,'' in
  \emph{Australasian Conference on Robotics and Automation}, 2018.

\bibitem{wang2019dob}
T.~Wang, W.~Lu, Z.~Yan, and D.~Liu, ``Dob-net: Actively rejecting unknown
  excessive time-varying disturbances,'' \emph{arXiv preprint
  arXiv:1907.04514}, 2019.

\bibitem{yu2017preparing}
W.~Yu, C.~K. Liu, and G.~Turk, ``Preparing for the unknown: Learning a
  universal policy with online system identification,'' in \emph{Proceedings of
  Robotics: Science and Systems}, Cambridge, Massachusetts, July 2017.

\bibitem{bengio2009curriculum}
Y.~Bengio, J.~Louradour, R.~Collobert, and J.~Weston, ``Curriculum learning,''
  in \emph{Proceedings of the 26th annual international conference on machine
  learning}, 2009, pp. 41--48.

\bibitem{caruana1997multitask}
R.~Caruana, ``Multitask learning,'' \emph{Machine learning}, vol.~28, no.~1,
  pp. 41--75, 1997.

\bibitem{taylor2009transfer}
M.~E. Taylor and P.~Stone, ``Transfer learning for reinforcement learning
  domains: A survey,'' \emph{Journal of Machine Learning Research}, vol.~10,
  no. Jul, pp. 1633--1685, 2009.

\bibitem{zhu2017target}
Y.~Zhu, R.~Mottaghi, E.~Kolve, J.~J. Lim, A.~Gupta, L.~Fei-Fei, and A.~Farhadi,
  ``Target-driven visual navigation in indoor scenes using deep reinforcement
  learning,'' in \emph{2017 IEEE international conference on robotics and
  automation (ICRA)}.\hskip 1em plus 0.5em minus 0.4em\relax IEEE, 2017, pp.
  3357--3364.

\bibitem{rusu2016progressive}
A.~A. Rusu, N.~C. Rabinowitz, G.~Desjardins, H.~Soyer, J.~Kirkpatrick,
  K.~Kavukcuoglu, R.~Pascanu, and R.~Hadsell, ``Progressive neural networks,''
  \emph{arXiv preprint arXiv:1606.04671}, 2016.

\bibitem{rusu2016sim}
A.~A. Rusu, M.~Vecerik, T.~Roth{\"o}rl, N.~Heess, R.~Pascanu, and R.~Hadsell,
  ``Sim-to-real robot learning from pixels with progressive nets,'' \emph{arXiv
  preprint arXiv:1610.04286}, 2016.

\bibitem{tzeng2020adapting}
E.~Tzeng, C.~Devin, J.~Hoffman, C.~Finn, P.~Abbeel, S.~Levine, K.~Saenko, and
  T.~Darrell, ``Adapting deep visuomotor representations with weak pairwise
  constraints,'' in \emph{Algorithmic Foundations of Robotics XII}.\hskip 1em
  plus 0.5em minus 0.4em\relax Springer, 2020, pp. 688--703.

\bibitem{bousmalis2018using}
K.~Bousmalis, A.~Irpan, P.~Wohlhart, Y.~Bai, M.~Kelcey, M.~Kalakrishnan,
  L.~Downs, J.~Ibarz, P.~Pastor, K.~Konolige, \emph{et~al.}, ``Using simulation
  and domain adaptation to improve efficiency of deep robotic grasping,'' in
  \emph{2018 IEEE International Conference on Robotics and Automation
  (ICRA)}.\hskip 1em plus 0.5em minus 0.4em\relax IEEE, 2018, pp. 4243--4250.

\bibitem{lennart1999system}
L.~Lennart, ``System identification: theory for the user,'' \emph{PTR Prentice
  Hall, Upper Saddle River, NJ}, pp. 1--14, 1999.

\bibitem{giri2010block}
F.~Giri and E.-W. Bai, \emph{Block-oriented nonlinear system
  identification}.\hskip 1em plus 0.5em minus 0.4em\relax Springer, 2010,
  vol.~1.

\bibitem{deisenroth2013survey}
M.~P. Deisenroth, G.~Neumann, J.~Peters, \emph{et~al.}, ``A survey on policy
  search for robotics,'' \emph{Foundations and Trends{\textregistered} in
  Robotics}, vol.~2, no. 1--2, pp. 1--142, 2013.

\bibitem{deisenroth2011pilco}
M.~Deisenroth and C.~E. Rasmussen, ``Pilco: A model-based and data-efficient
  approach to policy search,'' in \emph{Proceedings of the 28th International
  Conference on machine learning (ICML-11)}, 2011, pp. 465--472.

\bibitem{kuvayev1996model}
L.~Kuvayev and R.~S. Sutton, ``Model-based reinforcement learning with an
  approximate, learned model,'' in \emph{in Proceedings of the Ninth Yale
  Workshop on Adaptive and Learning Systems}.\hskip 1em plus 0.5em minus
  0.4em\relax Citeseer, 1996.

\bibitem{forbes2002representations}
J.~Forbes and D.~Andre, ``Representations for learning control policies,'' in
  \emph{Proceedings of the ICML-2002 Workshop on Development of
  Representations}, 2002, pp. 7--14.

\bibitem{hester2017intrinsically}
T.~Hester and P.~Stone, ``Intrinsically motivated model learning for developing
  curious robots,'' \emph{Artificial Intelligence}, vol. 247, pp. 170--186,
  2017.

\bibitem{jong2007model}
N.~K. Jong and P.~Stone, ``Model-based function approximation in reinforcement
  learning,'' in \emph{Proceedings of the 6th international joint conference on
  Autonomous agents and multiagent systems}.\hskip 1em plus 0.5em minus
  0.4em\relax ACM, 2007, p.~95.

\bibitem{sutton1991dyna}
R.~S. Sutton, ``Dyna, an integrated architecture for learning, planning, and
  reacting,'' \emph{ACM SIGART Bulletin}, vol.~2, no.~4, pp. 160--163, 1991.

\bibitem{brauer2012using}
C.~Brauer, ``Using eureqa in a stock day-trading application,'' \emph{Cypress
  Point Technologies, LLC}, 2012.

\bibitem{schmidt2009distilling}
M.~Schmidt and H.~Lipson, ``Distilling free-form natural laws from experimental
  data,'' \emph{science}, vol. 324, no. 5923, pp. 81--85, 2009.

\bibitem{abbeel2005exploration}
P.~Abbeel and A.~Y. Ng, ``Exploration and apprenticeship learning in
  reinforcement learning,'' in \emph{Proceedings of the 22nd international
  conference on Machine learning}, 2005, pp. 1--8.

\bibitem{gevers2006system}
M.~Gevers \emph{et~al.}, ``System identification without lennart ljung: what
  would have been different?'' \emph{Forever Ljung in System Identification,
  Studentlitteratur AB, Norrtalje}, vol.~2, 2006.

\bibitem{bongard2005nonlinear}
J.~C. Bongard and H.~Lipson, ``Nonlinear system identification using
  coevolution of models and tests,'' \emph{IEEE Transactions on Evolutionary
  Computation}, vol.~9, no.~4, pp. 361--384, 2005.

\bibitem{nagabandi2018neural}
A.~Nagabandi, G.~Kahn, R.~S. Fearing, and S.~Levine, ``Neural network dynamics
  for model-based deep reinforcement learning with model-free fine-tuning,'' in
  \emph{Robotics and Automation (ICRA), 2018 IEEE International Conference
  on}.\hskip 1em plus 0.5em minus 0.4em\relax IEEE, 2018, pp. 7579--7586.

\bibitem{peng2018sim}
X.~B. Peng, M.~Andrychowicz, W.~Zaremba, and P.~Abbeel, ``Sim-to-real transfer
  of robotic control with dynamics randomization,'' in \emph{2018 IEEE
  International Conference on Robotics and Automation (ICRA)}.\hskip 1em plus
  0.5em minus 0.4em\relax IEEE, 2018, pp. 1--8.

\bibitem{andrychowicz2020learning}
O.~M. Andrychowicz, B.~Baker, M.~Chociej, R.~Jozefowicz, B.~McGrew,
  J.~Pachocki, A.~Petron, M.~Plappert, G.~Powell, A.~Ray, \emph{et~al.},
  ``Learning dexterous in-hand manipulation,'' \emph{The International Journal
  of Robotics Research}, vol.~39, no.~1, pp. 3--20, 2020.

\bibitem{chebotar2019closing}
Y.~Chebotar, A.~Handa, V.~Makoviychuk, M.~Macklin, J.~Issac, N.~Ratliff, and
  D.~Fox, ``Closing the sim-to-real loop: Adapting simulation randomization
  with real world experience,'' in \emph{2019 International Conference on
  Robotics and Automation (ICRA)}.\hskip 1em plus 0.5em minus 0.4em\relax IEEE,
  2019, pp. 8973--8979.

\bibitem{tan2018sim}
J.~Tan, T.~Zhang, E.~Coumans, A.~Iscen, Y.~Bai, D.~Hafner, S.~Bohez, and
  V.~Vanhoucke, ``Sim-to-real: Learning agile locomotion for quadruped
  robots,'' \emph{arXiv preprint arXiv:1804.10332}, 2018.

\bibitem{lowrey2018reinforcement}
K.~Lowrey, S.~Kolev, J.~Dao, A.~Rajeswaran, and E.~Todorov, ``Reinforcement
  learning for non-prehensile manipulation: Transfer from simulation to
  physical system,'' in \emph{2018 IEEE International Conference on Simulation,
  Modeling, and Programming for Autonomous Robots (SIMPAR)}.\hskip 1em plus
  0.5em minus 0.4em\relax IEEE, 2018, pp. 35--42.

\bibitem{antonova2017reinforcement}
R.~Antonova, S.~Cruciani, C.~Smith, and D.~Kragic, ``Reinforcement learning for
  pivoting task,'' \emph{arXiv preprint arXiv:1703.00472}, 2017.

\bibitem{rajeswaran2016epopt}
A.~Rajeswaran, S.~Ghotra, B.~Ravindran, and S.~Levine, ``Epopt: Learning robust
  neural network policies using model ensembles,'' \emph{arXiv preprint
  arXiv:1610.01283}, 2016.

\bibitem{chiappa2017recurrent}
S.~Chiappa, S.~Racaniere, D.~Wierstra, and S.~Mohamed, ``Recurrent environment
  simulators,'' \emph{arXiv preprint arXiv:1704.02254}, 2017.

\bibitem{koryakovskiy2018model}
I.~Koryakovskiy, M.~Kudruss, H.~Vallery, R.~Babu{\v{s}}ka, and W.~Caarls,
  ``Model-plant mismatch compensation using reinforcement learning,''
  \emph{IEEE Robotics and Automation Letters}, vol.~3, no.~3, pp. 2471--2477,
  2018.

\bibitem{zarchan2013fundamentals}
P.~Zarchan and H.~Musoff, \emph{Fundamentals of Kalman filtering: a practical
  approach}.\hskip 1em plus 0.5em minus 0.4em\relax American Institute of
  Aeronautics and Astronautics, Inc., 2013.

\end{thebibliography}
\end{document}